\documentclass[letterpaper]{article} % DO NOT CHANGE THIS
\usepackage{aaai24}  % DO NOT CHANGE THIS
\usepackage{times}  % DO NOT CHANGE THIS
\usepackage{helvet}  % DO NOT CHANGE THIS
\usepackage{courier}  % DO NOT CHANGE THIS
\usepackage[hyphens]{url}  % DO NOT CHANGE THIS
\usepackage{graphicx} % DO NOT CHANGE THIS
\usepackage{natbib}  % DO NOT CHANGE THIS AND DO NOT ADD ANY OPTIONS TO IT
\usepackage{caption} % DO NOT CHANGE THIS AND DO NOT ADD ANY OPTIONS TO IT
\usepackage{algorithm}
\usepackage{algorithmic}

\usepackage{newfloat}
\usepackage{listings}
\DeclareCaptionStyle{ruled}{labelfont=normalfont,labelsep=colon,strut=off} % DO NOT CHANGE THIS
\floatstyle{ruled}
\newfloat{listing}{tb}{lst}{}
\floatname{listing}{Listing}
\nocopyright 

\title{Closing the Gap: Achieving Better Accuracy-Robustness Tradeoffs against Query-Based Attacks}
\author{
    Pascal Zimmer\textsuperscript{\rm 1}, Sébastien Andreina\textsuperscript{\rm 2}, Giorgia Azzurra Marson\textsuperscript{\rm 2}, Ghassan Karame\textsuperscript{\rm 1}
}
\affiliations{
    \textsuperscript{\rm 1}Ruhr-Universität Bochum, Germany\\
    \textsuperscript{\rm 2}NEC Labs Europe, Germany\\
    \{pascal.zimmer, ghassan.karame\}@rub.de, \{sebastien.andreina, giorgia.marson\}@neclab.eu    
}

\usepackage{subcaption}
\usepackage{booktabs}
\usepackage{nicefrac}

\usepackage{amsmath}
\usepackage{amssymb}
\usepackage[capitalize]{cleveref}

\usepackage{multirow}

\DeclareMathOperator*{\argmax}{arg\,max}

\newcommand{\RR}{\mathbb{R}}
\newcommand{\size}[1]{\lvert #1 \rvert} % cardinality of sets

\newcommand{\setX}{\mathcal{X}}
\newcommand{\setY}{\mathcal{Y}}

\newcommand{\adv}{\mathcal{A}} % adversary
\newcommand{\dist}[1]{\lVert #1 \rVert} % distance metric
\newcommand{\Classifier}{C} % Generic symbol for a classifier (with or without defense)

\newcommand{\advsuccrate}{\mathsf{ASR}}
\newcommand{\robustacc}{\mathsf{RA}}
\newcommand{\mainacc}{\mathsf{CA}}
\newcommand{\defense}{\mathsf{D}}
\newcommand{\rnd}{\mathsf{RND}}
\newcommand{\rcr}{\mathsf{RCR}}
\newcommand{\jpeg}{\mathsf{JPEG}}

\usepackage{array}
\newcolumntype{H}{>{\setbox0=\hbox\bgroup}c<{\egroup}@{}}

\newtheorem{definition}{Definition}

\newtheorem{proposition}{Proposition}

\begin{document}

\maketitle

\begin{abstract}
    \hyphenation{Image-Net}

    Although promising, existing defenses against query-based attacks share a common limitation: they offer increased robustness against attacks at the price of a considerable accuracy drop on clean samples.
    In this work, we show how to efficiently establish, at test-time, a solid tradeoff between robustness and accuracy when mitigating query-based attacks.
    Given that these attacks necessarily explore low-confidence regions,
    our insight is that activating dedicated defenses, such as random noise defense and random image transformations, only for low-confidence inputs is sufficient to prevent them.
    Our approach is independent of training and supported by theory.
    We verify the effectiveness of our approach for various existing defenses by conducting extensive experiments on CIFAR-10, CIFAR-100, and ImageNet. Our results confirm that our proposal can indeed enhance these defenses by providing better tradeoffs between robustness and accuracy when compared to state-of-the-art approaches while being completely training-free\footnote{The corresponding code is available at \url{https://github.com/RUB-InfSec/closing_the_gap}.}.
\end{abstract}

\section{Introduction}

Even though deep neural networks (DNNs) are currently enjoying broad applicability, they are unfortunately fragile to (easily realizable) manipulations of their inputs. Namely, \textit{adversarial samples} pose a severe threat to the deployment of DNNs in safety-critical applications, e.g., autonomous driving and facial recognition. For instance, in the context of image classification, adversarial samples can deceive a  classifier with carefully crafted and visually almost imperceptible perturbations applied to an input image. This results in (un-)targeted misclassification with identical semantic information to the human eye.

Initially, all attack strategies were designed in the so-called white-box model~\cite{biggioEvasionAttacksMachine2013, goodfellowExplainingHarnessingAdversarial2015}, in which the attacker has full domain knowledge, e.g., model architecture, trained parameters, and training data. More recent attacks, such as query-based attacks, consider a better grounded and more realistic threat model, in which the attacker has no knowledge about the classifier's internals and training data and is only able to observe outputs to supplied inputs, i.e., through oracle access to the classifier. This black-box setting faithfully mimics many real-world applications, such as existing machine learning as a service (MLaaS) deployments.
Query-based black-box attacks can be categorized as \emph{score-based} or \emph{decision-based} attacks~\cite{chenHopSkipJumpAttackQueryEfficientDecisionBased2020a, mahoSurFreeFastSurrogatefree2021a}. The former category assumes that the adversary can acquire information from the output score of the classifier, while the latter mimics a more realistic setting where the adversary has only access to the top-1 label of the classifier's prediction.

In an attempt to design defensive strategies to mitigate adversarial inputs, an arms race has been sparked in the community. A number of proposals have explored the use of randomization to improve adversarial robustness.
While most randomization strategies are ineffective in the original white-box setting~\cite{athalyeSynthesizingRobustAdversarial, athalyeObfuscatedGradientsGive2018}, recent findings suggest that embedding random noise within the input could effectively mitigate query-based black-box attacks~\cite{byunEffectivenessSmallInput2022}. As such, while randomization-based defenses could be effective in thwarting query-based attacks, they inevitably damage the classifier's clean accuracy \cite{tsiprasRobustnessMayBe2019}.
{For example, the random noise defense~\cite{qinRandomNoiseDefense} (using the noise level $\sigma=0.07$ as a hyperparameter) can increase the robust accuracy against PopSkipJump (PSJA)~\cite{simon-gabrielPopSkipJumpDecisionBasedAttack2021a}, the strongest known black-box attack against randomized classifiers, by almost $13\%$ at the cost of a significant drop of almost $30\%$ in main task accuracy in the CIFAR-10 dataset.}

In this work, we set forth to establish a stronger accuracy-robustness tradeoff against query-based black-box attacks by leveraging a different hyperparameter grounded on the confidence $\tau$ of classifying incoming inputs. Our approach relies on the insight that, while query-based attacks necessarily need to explore low-confidence regions, most genuine inputs are classified with relatively high confidence.
By cleverly differentiating between low- and high-confidence regions, we aim to establish a strong tradeoff between adversarial robustness and the model's clean accuracy. To do so, we propose to obstruct the search for adversarial inputs by activating a defensive layer (e.g., based on input randomization) \emph{only for low-confidence inputs}.
That is, we propose to only activate (existing) randomization defenses on inputs~$x$ such that~{$\max_i f_i(x) < \tau$}, for an appropriate threshold~$0\leq \tau \leq 1$, while high-confidence inputs (with confidence at least~$\tau$) are processed normally.

We show that our approach can be instantiated with existing test-time defenses without the need for retraining and show that it can strike robust tradeoffs that could not be reached otherwise with existing off-the-shelf defenses.
While our approach is generic, it is naturally apt to particularly thwart decision-based attacks, as it is harder for the adversary to avoid low-confidence regions when she does not have access to the scores output by the classifier.

We conduct an extensive robustness evaluation of our approach with lightweight off-the-shelf defenses and find across-the-board improvements in accuracy-robustness tradeoffs over all considered defenses. For instance, our experiments on CIFAR-10 and CIFAR-100 show that for PSJA, one of the most powerful state-of-the-art decision-based attacks, our method improves robust accuracy by up to~{$9\%$ and $20\%$}, with a negligible impact on clean task accuracy of at most $2\%$.
For SurFree, a geometric decision-based attack, we report robustness improvements of up to {$34\%$} with almost no degradation of main task accuracy.

\section{Related Work}
\label{section:related_work}

\paragraph{Black-box attacks.}

In contrast to white-box attackers, who can easily generate adversarial samples using gradients of the model, a black-box attacker is unaware of the classifier's internals~\cite{DBLP:conf/ccs/SharadMTK20}. In a black-box attack, the adversary only accesses the target classifier as an oracle and uses its responses to generate adversarial examples.

Transfer-based attacks~\cite{Dong_2018_CVPR, DBLP:conf/cvpr/XieZZBWRY19, ttp} have access to (similar) training data that has been used to train the target classifier. Using such datasets, they train a local ``surrogate'' model and use it to craft adversarial samples with white-box strategies against the surrogate model. Due to the transferability property of DNNs, the generated samples often fool the original target classifier. To be effective, these attacks require knowledge of the target classifier's architecture and/or its training data.
On the other hand, query-based attacks~\cite{brendelDecisionbasedAdversarialAttacks2018a,square_attack, chenHopSkipJumpAttackQueryEfficientDecisionBased2020a} do not require access to the training data itself and instead interact with the target classifier to obtain predictions on inputs of their choice.
By adaptively generating a sequence of images based on the classifier's predictions, query-based attacks can derive adversarial examples with minimal distortion.
Score-based attacks assume additional access to the individual probabilities of the possible classes, while \emph{decision-based} attacks typically work with access to the top-1 label.

\paragraph{Decision-based attacks.} \label{sec:rw:decision_based}

\begin{figure}[tb]
    \center{
        \includegraphics[width=1.0\columnwidth]{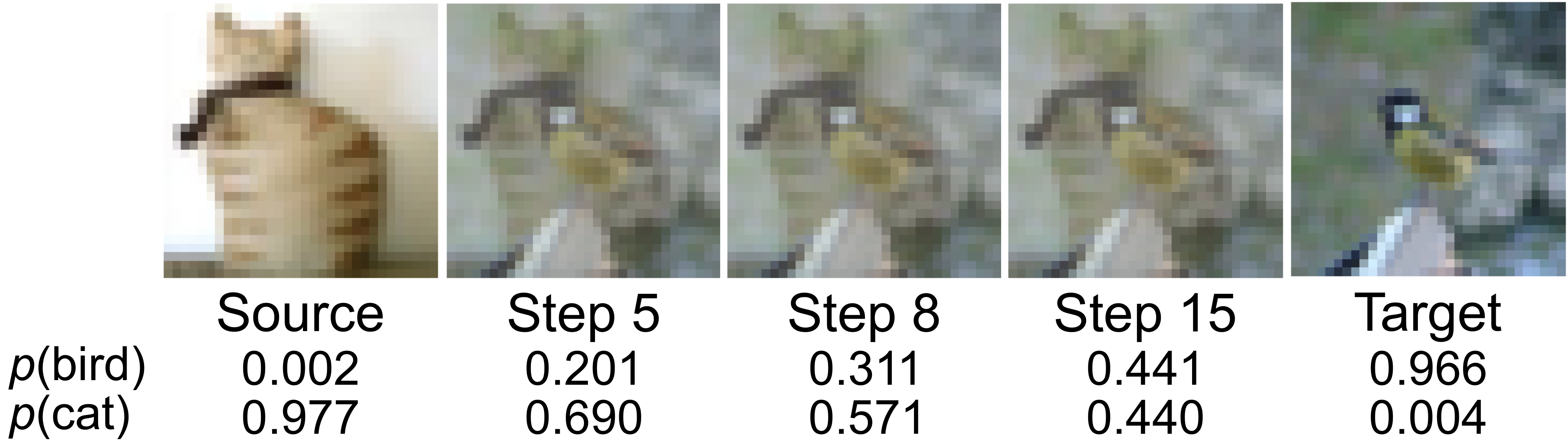}
    }
    \caption{Selected iterations (or steps) from the binary search procedure typically used in a decision-based attack.
        Here, a source image, classified as 'cat', is blended with a target image, classified as 'bird'. The procedure crosses a low-confidence region before outputting a boundary sample with a slightly higher probability for the target class.}
    \label{fig:binary_search}
\end{figure}

As decision-based attacks can only obtain the top-1 label of a prediction, they cannot leverage the classifier's confidence.

Hence they attempt to locate the classifier's decision boundary, as seen in \Cref{fig:binary_search}, and to estimate the shape or the gradient near the boundary or by exploiting its geometric properties. The ultimate goal of such attacks is to undergo a series of steps with the aim of gradually minimizing the distortion of each candidate adversarial sample.

The first known attack~\cite{brendelDecisionbasedAdversarialAttacks2018a} is based on a naive rejection-sampling exploration of the decision boundary to converge to a low-distortion adversarial example, which requires a significant amount of queries.
Recent proposals can primarily be divided into two broader categories: gradient-based methods, such as HopSkipJump (HSJA)~\cite{chenHopSkipJumpAttackQueryEfficientDecisionBased2020a}, build a surrogate gradient inspired by zeroth-order-optimization for faster convergence, and geometry-based attacks, such as SurFree~\cite{mahoSurFreeFastSurrogatefree2021a}, exploit geometric properties of the decision boundary and perform careful query trials along it.

Since decision-based attacks are based on the assumption of a deterministic classifier, they are especially susceptible to input-randomization defenses~\cite{chenHopSkipJumpAttackQueryEfficientDecisionBased2020a, qinRandomNoiseDefense}. By adapting HSJA to noisy environments, PSJA~\cite{simon-gabrielPopSkipJumpDecisionBasedAttack2021a} overcomes this fragility.

\paragraph{Defenses.} \label{sec:rw:defenses}

Existing mitigation strategies against adversarial examples can be categorized into two main groups: training- and inference-time approaches.
The most popular training-time defense is adversarial training with various instantiations~\cite{madryDeepLearningModels2019, zhangTheoreticallyPrincipledTradeoff}, which essentially augments the training dataset with adversarial examples labeled with the semantically correct class. This process is especially costly as the adversarial images have to be generated at training time to allow their use during the training procedure.  {Other approaches, such as Parametric Noise Injection (PNI) \cite{he2019parametric}, inject Gaussian distributed noise upon weights in a layer-wise fashion or insert a noise layer before each convolution layer, as done in Random Self Ensemble (RSE)~\cite{liu2018towards}.}

Inference-time defenses are generally more lightweight. Popular examples include randomized pre-processing strategies such as random noise defense (RND)~\cite{qinRandomNoiseDefense}, random resize and cropping (RCR)~\cite{guoCounteringAdversarialImages2018a,xieMitigatingAdversarialEffects2018d}.
These defenses promise increased robustness against decision-based attacks and come at a small additional inference cost. Deterministic inference-time defenses include applying the JPEG compression algorithm and have been shown to enhance the robustness of a classifier~\cite{guoCounteringAdversarialImages2018a, dziugaiteStudyEffectJPG2016}.

However, all these defenses share a common limitation: they affect benign samples similarly to adversarial examples, hence inevitably reducing main task accuracy. For instance, RND requires a noise level below $\sigma=0.02$ in order to preserve main task accuracy, while $\sigma = 0.05$ is necessary to reach a reasonable level of robustness above $70\%$ for CIFAR-10 and PSJA.

\paragraph{Accuracy-Robustness Tradeoff.}
There is already great progress in analyzing existing tradeoffs between accuracy and robustness~\cite{zhangTheoreticallyPrincipledTradeoff, stutzDisentanglingAdversarialRobustness2019, schmidtAdversariallyRobustGeneralization, tsiprasRobustnessMayBe2019}. For instance, recent findings show that robust and accurate classifiers are possible for certain (realistic) classification tasks, as long as different classes are sufficiently separated~\cite{yangCloserLookAccuracy2020}.

Most proposals that aim at improving the accuracy-robustness tradeoff are focused on adversarial-training defenses based on data-augmentation methods, hence leading to high computational overhead at training time~\cite{raghunathanUnderstandingMitigatingTradeoff2020}.
{Other approaches allow for a ``free'' adjustment with a hyperparameter, without requiring any retraining after initially augmenting the model with a new model-conditional training approach~\cite{wangOnceforallAdversarialTraining2020}.}

\section{Methodology}

\subsection{Preliminaries and Notations}

We define an adversarial sample $x'$ as a genuine image $x$ to which carefully crafted adversarial noise $p$ is added, i.e., $x' = x + p$ for a small perturbation~$p$ such that~$x'$ and~$x$ are perceptually indistinguishable to the human eye and yet are classified differently.

Let~$f\colon \RR^d\to \Delta^n$ be a DNN model assigning $d$-dimensional inputs to~$n$ classes, where~$\Delta^n$ is the probability vector of~$n$ classes, and let~$\Classifier\colon \RR^d\to [n]$ be the associated classifier defined as $\Classifier(x) := \argmax_{i \in [n]} f_i(x)$. The highest prediction probability $\max_{i\in[n]}f_i(x)$ is called the classifier's \emph{confidence}.

Given a genuine input~$x_0\in \RR^d$ predicted as~${C(x_0) = s}$ (source class), $x'$ is an \emph{adversarial sample} of~$x_0$ if 
$C(x')~\neq~s$ and
$\dist{x'-x_0}_p \leq \varepsilon$ for a given distortion bound~$\varepsilon\in \RR^+$ and $l_p$ norm. Formally, we have:
\begin{equation}
    \label{eq:misclassifications}
    \begin{cases}
        C(x') \neq s & \text{(untargeted attack),} \\
        C(x') = t    & \text{(targeted attack).}
    \end{cases}
\end{equation}
The objective of the adversary can then be expressed as follows:
\begin{equation}
    \label{eq:adversarial:objective}
    \adv_{x_0}(x) :=
    \begin{cases}
        \max\limits_{i\neq s} f_i(x) - f_{s}(x) & \text{(untargeted attack),} \\
        f_t(x) - \max\limits_{i\neq t}f_i(x)    & \text{(targeted attack).}
    \end{cases}
\end{equation}
{For an adversarial sample~$x'$ to be successful, it must have a low distortion and satisfy~$\adv_{x_0}(x') > 0$. The attacker searches for inputs~$x'$ solving the optimization problem, while she is restricted by the query budget $Q$ in her number of permitted queries to the classifier:
\begin{equation}
    \min_{x'} \dist{x' - x_0}_p
    \quad\text{such that}\quad
    \adv_{x_0}(x') > 0.
\end{equation}}%
In contrast to score-based attacks---that receive the classifier's scores~$(f_i(x))_{i\in[n]}$ for each queried input~$x$ and can thus evaluate the function~$\adv_{x_0}(x)$---decision-based attacks only obtain the final prediction~$\argmax_i f_i(x)$ and cannot compute~$\adv_{x_0}(x)$.
Nevertheless, the prediction~$C(x')$ is sufficient to determine the sign of~$\adv_{x_0}(x')$ as per~Equation~\eqref{eq:adversarial:objective}.

\subsection{Main Intuition}\label{subsec:approach}

\begin{figure}[tb]
    \center{
        \includegraphics[width=0.80\columnwidth]{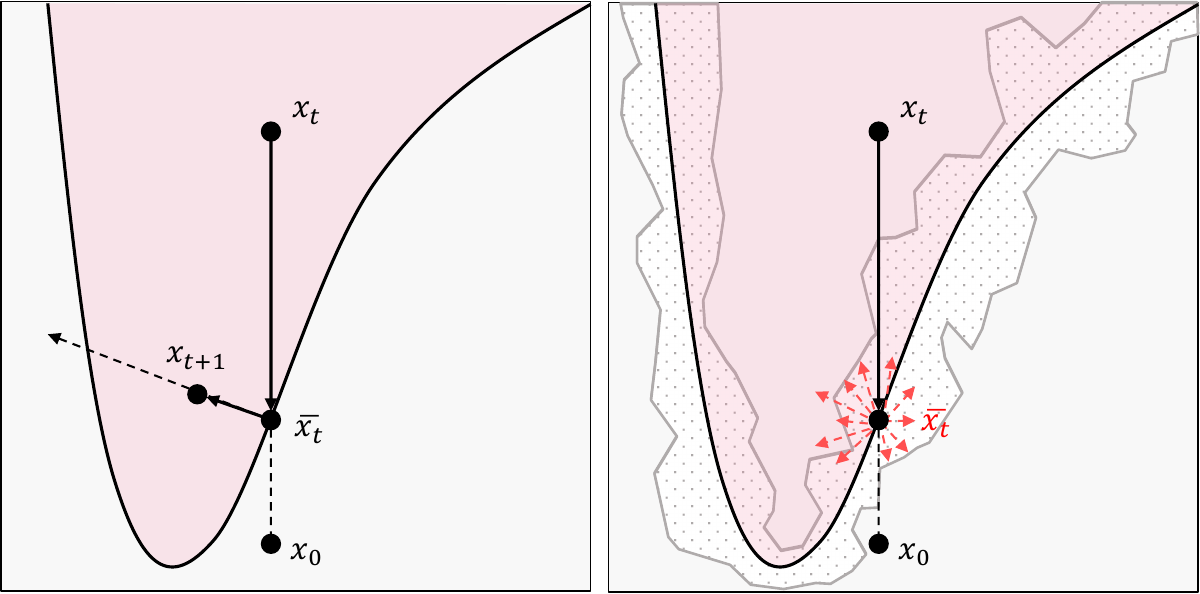}
    }
    \caption{Iteration of a decision-based attack for an unprotected classifier (left) and in the presence of our approach (right). The attacker starts with a sample~$x_t$ in the target class, and in each iteration, she updates the current adversarial sample, from~$x_t$ to~$x_{t+1}$, by locating the boundary sample~$\bar{x}_t$.
        Intuitively, activating the defense only in low-confidence regions (right) can effectively obstruct the search for adversarial perturbations.
    }
    \label{fig:boundary:samples}
\end{figure}

We propose an approach that specifically aims to impede the search for adversarial perturbations in low-confidence regions while optimistically treating high-confidence samples as non-adversarial. A central component of decision-based attacks is a binary search procedure that is used to locate the decision boundary in a low-confidence region for further exploration.

More specifically, we propose to dynamically trigger the activation of a suitable defensive layer depending on the classifier's confidence on its input. Here, we consider lightweight defenses, such as RND, RCR, and JPEG.

Formally, our method is parameterized by a \emph{confidence threshold} $\tau \in [0,1]$ as a hyperparameter and can be generically combined with any defensive technique~$\defense$ applied at inference time.
\begin{definition}[Low-confidence region]
    We define a \emph{low-confidence region} with respect to classifier $f$ and confidence threshold $\tau$ as the space of samples for which~$f$ has confidence below $\tau$, i.e., $\{ x\in \mathcal{X} : \max_{i\in[n]} f_i(x) < \tau\}$.
\end{definition}
Namely, let~$f$ and~$f_\defense$ denote the (unprotected) model and its protected version with defense~$\defense$, respectively.
Our strategy consists of invoking the defense only when the classifier has confidence below the threshold, else the unprotected classifier is used, resulting in the following classifier:
\begin{equation}
    F_{\defense,\tau}(x) :=
    \begin{cases}
        f_\defense(x) & \text{if }\max_{i\in[n]} f_i(x) < \tau, \\
        f(x)          & \text{otherwise.}
    \end{cases}
\end{equation}
Due to the limited information retrieved in the decision-based model, attacks usually start with a highly distorted image (i.e, $x_1=x+p$, $p>>\varepsilon$) such that $\adv_{x_0}(x_1)>0$ as a starting point. Afterward, it follows the iterative method displayed in Figure~\ref{fig:boundary:samples} (left) where it first locates the point $\bar{x_t}$  at the decision boundary through a binary search between $x_0$ and $x_t$, before pursuing further exploration strategies. An example is a gradient estimation to minimize the distortion and generate $x_{t+1}$.

Our intuition is that obstructing the generation of low-confidence adversarial inputs is sufficient to thwart the exact location of $\bar{x_t}$ and the further exploration of the decision boundary at that point as illustrated in Figure~\ref{fig:boundary:samples} (right).

As most genuine samples are classified correctly with high confidence, the defense has minimal impact on the classification accuracy of genuine samples while still being effective at preventing attacks, as attackers have to visit the low-confidence region where the defense is enabled in order to create adversarial samples.

\subsubsection{Calibration.}

Predictions of recent model architectures are typically miscalibrated in the sense that they usually classify inputs with rather high confidence. To make our approach agnostic to the deployment instance, we calibrated our models by leveraging temperature scaling~\cite{guoCalibrationModernNeural}, i.e., using a single parameter $T>0$ to scale the logits $z$ before passing them through softmax $\sigma$ and updating the confidence prediction $p$ with $p=\sigma(\nicefrac{z}{T})$.

The parameter $T$ is computed so as to minimize the difference between the reported confidence (i.e., the value $p$) and the actual accuracy. We group the predictions into $M$ interval bins and denote the set of images falling in the \emph{m}-th bin as $B_m$. Then, we derive~$T$ by solving the optimization problem below:
\begin{equation}
    \min_T \sum_{m=1}^{M}\frac{|B_m|}{n}\left| \frac{1}{|B_m|}\sum_{i\in B_m}\textbf{1}(\hat{y}_i=y_i) - \frac{1}{|B_m|}\sum_{i\in B_m}\hat{p_i}\right|
\end{equation}
\noindent where $n$ is the overall number of samples, $y_i$ the true class label, $\hat{y}_i$ the predicted class label, and $\hat{p}_i$ the confidence of the model when classifying image $i$. 
The indicator function $\mathbf{1}$ is defined as:
\[
    \mathbf{1}(x):=\begin{cases}
        1\;\mathrm{if}\;x\;\mathrm{is\;true} \\
        0\;\mathrm{otherwise.}               \\
    \end{cases}
\]
We stress that our approach is independent of model calibration  (i.e., low-confidence regions must still be explored by attacks). Namely, we merely expect that the threshold $\tau$ in our approach to slightly vary among calibrated and uncalibrated models---with no impact on the resulting accuracy or robustness.

\subsubsection{Theoretical Motivation.}

Recall that the supervised training procedure for a generic multi-class classifier $C$ aims at finding the optimal set of parameters $\theta$ that minimize the aggregated loss over the entire training set $\mathcal{X}$: $
    \min_{\theta}\frac{1}{N}\sum_{i=1}^{N}\mathcal{L}(y_i, f(x_i, \theta))$,
for $N=|\mathcal{X}|$, sample $x$, ground-truth $y$, and loss-function $\mathcal{L}$.

\noindent The optimization procedure itself can be instantiated with varying algorithms, such as SGD, which come with varying convergence guarantees. Depending on the use case, loss-function $\mathcal{L}$ is, for example, the cross-entropy loss.

In what follows, we show that \emph{(1)} genuine (i.e., non-adversarial) samples are unlikely to bear low-confidence, and \emph{(2)} existing query-based attacks necessarily have to investigate low-confidence regions to ensure convergence.

\begin{proposition}
    A set of parameters $\theta$ trained on $\mathcal{X}$ are likely to classify images following this distribution with high confidence.
\end{proposition}
Consider the loss function $\mathcal{L}$ with the $n$-class cross-entropy loss:
\begin{equation}
    \mathcal{L}(y, \mathbf{p}) := -\sum_{c=1}^{n}\mathbf{1}(y = c)\log(p_c)
\end{equation}
with $\mathbf{p}$ being the probability vector output by the classifier. 

Now, let us define $q:=\mathbf{1}(y=c)$ as a fixed reference probability distribution, i.e., the one-hot encoding of $y$. Due to Gibbs' inequality~\cite{book112} the cross-entropy function takes its minimum when $p=q$, i.e., the network is optimizing towards predictions of high confidence. Notice that this result corroborates findings in~\cite{guoCalibrationModernNeural,devries2018learning} which also suggest that models tend to output high confidence predictions for samples following the training distribution $\mathcal{X}$. An empirical validation of Proposition 1 can be found in the Appendix.

\begin{proposition}
    During the binary-search procedure of a decision-based attack,  $\max_{i\in[n]}f_i(x) \leq \nicefrac{1}{2}$.
\end{proposition}
Decision-based attacks, such as PSJA~\cite{simon-gabrielPopSkipJumpDecisionBasedAttack2021a} and SurFree~\cite{mahoSurFreeFastSurrogatefree2021a}, use the classifier's feedback on input queries in order to locate the \emph{boundary}:
\begin{equation}
    \label{eq:boundary}
    \mathrm{bd}(\adv_{x_0}) := \{ x \in \RR^d : \adv_{x_0}(x) = 0\}
\end{equation}
A binary-search computes an interpolation $x$ between a source image $x_0$ and a target image $x_t$, satisfying~${\adv_{x_0}(x_t) > 0}$ -- more precisely:
\begin{equation} x:= \mathrm{bs}(x_t, x_0, k) = (1-k)x_t + kx_0 \end{equation}
with interpolation factor $k \in [0, 1]$.

Figure~\ref{fig:boundary:samples} (left) illustrates one iteration of a binary search (and gradient sampling); here, each iteration reduces the distance from the candidate sample~$x_t$ and the original sample~$x_0$. Therefore, the attacker needs to query the classifier on intermediate samples~$\bar{x}_t$ that lie on the boundary, i.e., \emph{where the classifier's confidence is low} in order to estimate the gradient or the geometric shape of the boundary.

The proof for Proposition 2 can be found in the Appendix.

\section{Experiments}

In this section, we empirically evaluate our approach on the CIFAR-10, CIFAR-100, and ImageNet datasets and compare its efficacy to related work in the area.

\paragraph{Datasets.} In line with the literature, we evaluate our approach on datasets of varying input dimensions and number of classes, i.e., CIFAR-10, CIFAR-100~\cite{krizhevskyLearningMultipleLayers2009}, and ImageNet~\cite{ILSVRC15} datasets. The former two contain $50,000$ train and $10,000$ test images of size $32 \times 32$ pixels, divided into 10 and 100 different classes, respectively. The latter contains $1.2$ million training images and a validation set of $50,000$ images.

\paragraph{Attack choice.}
Our choice of attacks is mainly motivated by a study that provides an overview of decision-based attacks and outlines a comparison of their contributions. The study can be found in the Appendix. We selected PSJA~\cite{simon-gabrielPopSkipJumpDecisionBasedAttack2021a} as it is the only attack that considers a probabilistic classifier. It is based on gradient estimation and is a direct improvement over HSJA~\cite{chenHopSkipJumpAttackQueryEfficientDecisionBased2020a}.
We further selected the SurFree attack~\cite{mahoSurFreeFastSurrogatefree2021a} as it is the most recent attack, which trades the costly gradient estimation step and instead exploits geometric properties of the decision boundary.
We argue that our selected attacks exhibit good coverage over attacks in the literature.

\paragraph{Defense choice.}
Our approach can be generically instantiated using any inference-time defense~$\defense$.
We selected three lightweight transformations to instantiate our approach: RND and RCR for the probabilistic setting and JPEG compression for the deterministic setting.

RND applies additive centered Gaussian noise to the input
$
    f_\rnd(\nu,x) = f(x + \nu r)
    \quad\text{for}\quad
    r \gets \mathcal{N}(0,I),
$
where the parameter~$\nu\in\RR$ controls the noise magnitude.

RCR is a function $\gamma_\nu$ that randomly crops and bi-linearly interpolates the image back to its original size:
$
    {f_\rcr(\nu,x) = f(\gamma_\nu(x))},
$ where~$\nu$ denotes the cropping size.

Finally, JPEG applies the JPEG compression algorithm $\phi$ with
$	f_\jpeg(\nu,x) = f(\phi_\nu(x)),$where $\nu\in [0,100\%]$ is the quality parameter.
We include a deterministic defense to evaluate attacks, such as SurFree, that rely on deterministic classification, as they are otherwise easily defeated by probabilistic defenses~\cite{chenHopSkipJumpAttackQueryEfficientDecisionBased2020a}.

\newcommand{\textnf}[1]{\textrm{#1}}
\newcommand{\shp}[1]{\textbf{\shortstack{#1}}}
\newcommand{\shs}[1]{\shortstack{#1}}

\begin{table*}[t]
    \centering
    \footnotesize
        \begin{tabular}{cc|c|HHHccccHccc|HHHcccccccc}
            \toprule
             & Dataset        &                 & \multicolumn{11}{c|}{CIFAR-10} & \multicolumn{11}{c}{CIFAR-100}                                                                                                                                                                                         \\
             & $\nu$ / $\tau$ &                 & 0.0                            & 0.3                             & 0.4                           & 0.5 & 0.6 & 0.7 & 0.8 & 0.9 & 0.97 & 0.99 & 1.0 & 0.0 & 0.3 & 0.4 & 0.5 & 0.6 & 0.7 & 0.8 & 0.9 & 0.97 & 0.99 & 1.0 \\
            \midrule

            \parbox[t]{2mm}{\multirow{5}{*}{\rotatebox[origin=c]{90}{\shortstack{PSJA\\ with RND}}}}
& \shs{0.02} &\shs{$\mainacc$\\$\robustacc$} &\shs{0.95\\0.47}&\shs{0.95\\0.42}&\shs{0.95\\0.46}&\shp{0.95\\0.51}&\shp{0.95\\0.54}&\shp{0.94\\0.57}&\shp{0.94\\0.57}&\shp{0.94\\0.56}&\shp{0.94\\0.62}&\shp{0.94\\0.55}&\shs{0.94\\0.53}&\shs{0.60\\0.15}&\shs{0.60\\0.17}&\shs{0.60\\0.18}&\shp{0.59\\0.26}&\shp{0.59\\0.29}&\shp{0.59\\0.28}&\shp{0.59\\0.33}&\shp{0.59\\0.31}&\shp{0.58\\0.34}&\shp{0.58\\0.33}&\shs{0.58\\0.26}\\\addlinespace
& \shs{0.05} &\shs{$\mainacc$\\$\robustacc$} &\shs{0.95\\0.47}&\shp{0.95\\0.49}&\shs{0.95\\0.43}&\shs{0.95\\0.44}&\shs{0.94\\0.49}&\shs{0.94\\0.47}&\shp{0.94\\0.59}&\shp{0.93\\0.54}&\shp{0.87\\0.64}&\shp{0.83\\0.71}&\shs{0.82\\0.71}&\shs{0.60\\0.15}&\shs{0.59\\0.14}&\shs{0.59\\0.14}&\shs{0.58\\0.23}&\shs{0.56\\0.26}&\shp{0.55\\0.31}&\shp{0.53\\0.38}&\shs{0.50\\0.33}&\shs{0.48\\0.38}&\shs{0.47\\0.40}&\shs{0.46\\0.46}\\\addlinespace
& \shs{0.07} &\shs{$\mainacc$\\$\robustacc$} &\shs{0.95\\0.47}&\shs{0.95\\0.47}&\shs{0.95\\0.45}&\shs{0.95\\0.44}&\shp{0.94\\0.54}&\shs{0.94\\0.48}&\shs{0.93\\0.51}&\shp{0.92\\0.54}&\shs{0.82\\0.54}&\shs{0.65\\0.58}&\shs{0.65\\0.60}&\shs{0.60\\0.15}&\shs{0.59\\0.13}&\shs{0.58\\0.13}&\shs{0.56\\0.22}&\shs{0.54\\0.20}&\shs{0.53\\0.28}&\shs{0.49\\0.35}&\shs{0.46\\0.37}&\shs{0.41\\0.37}&\shs{0.38\\0.44}&\shs{0.37\\0.37}\\
\midrule\parbox[t]{2mm}{\multirow{11}{*}{\rotatebox[origin=c]{90}{PSJA with RCR}}}
& \shs{29\\200} &\shs{$\mainacc$\\$\robustacc$} &\shs{0.95\\0.41}&\shp{0.95\\0.44}&\shp{0.95\\0.42}&\shs{0.95\\0.37}&\shp{0.95\\0.47}&\shp{0.94\\0.44}&\shs{0.94\\0.35}&\shp{0.94\\0.42}&\shs{0.92\\0.32}&\shs{0.91\\0.33}&\shs{0.91\\0.37}&\shs{0.60\\0.17}&\shs{0.60\\0.16}&\shp{0.60\\0.18}&\shp{0.59\\0.24}&\shp{0.59\\0.29}&\shp{0.58\\0.31}&\shp{0.56\\0.37}&\shp{0.55\\0.42}&\shp{0.54\\0.36}&\shp{0.54\\0.44}&\shs{0.53\\0.35}\\\addlinespace
& \shs{27\\176} &\shs{$\mainacc$\\$\robustacc$} &\shs{0.95\\0.41}&\shs{0.95\\0.41}&\shs{0.95\\0.39}&\shp{0.95\\0.43}&\shs{0.95\\0.39}&\shp{0.94\\0.41}&\shp{0.94\\0.41}&\shs{0.94\\0.38}&\shp{0.90\\0.42}&\shs{0.88\\0.39}&\shs{0.88\\0.34}&\shs{0.60\\0.17}&\shs{0.60\\0.15}&\shs{0.59\\0.17}&\shp{0.59\\0.25}&\shp{0.58\\0.37}&\shp{0.57\\0.40}&\shp{0.56\\0.42}&\shp{0.53\\0.43}&\shs{0.51\\0.41}&\shp{0.50\\0.49}&\shs{0.50\\0.46}\\\addlinespace
& \shs{26\\152} &\shs{$\mainacc$\\$\robustacc$} &\shs{0.95\\0.41}&\shp{0.95\\0.47}&\shs{0.95\\0.39}&\shs{0.95\\0.38}&\shs{0.95\\0.41}&\shp{0.94\\0.45}&\shs{0.94\\0.39}&\shp{0.93\\0.47}&\shs{0.89\\0.33}&\shs{0.86\\0.34}&\shs{0.86\\0.39}&\shs{0.60\\0.17}&\shs{0.60\\0.13}&\shp{0.59\\0.21}&\shs{0.58\\0.21}&\shp{0.58\\0.37}&\shp{0.56\\0.36}&\shp{0.54\\0.39}&\shp{0.52\\0.49}&\shp{0.49\\0.47}&\shp{0.47\\0.50}&\shs{0.47\\0.48}\\\addlinespace
& \shs{25\\128} &\shs{$\mainacc$\\$\robustacc$} &\shs{0.95\\0.41}&\shp{0.95\\0.46}&\shs{0.95\\0.40}&\shs{0.95\\0.41}&\shs{0.95\\0.35}&\shp{0.94\\0.43}&\shs{0.94\\0.39}&\shs{0.93\\0.35}&\shp{0.88\\0.46}&\shs{0.83\\0.34}&\shs{0.83\\0.33}&\shs{0.60\\0.17}&\shs{0.60\\0.15}&\shp{0.59\\0.21}&\shp{0.58\\0.27}&\shp{0.57\\0.28}&\shp{0.56\\0.34}&\shp{0.54\\0.44}&\shs{0.50\\0.42}&\shs{0.47\\0.45}&\shs{0.45\\0.47}&\shs{0.44\\0.47}\\\addlinespace
& \shs{22\\104} &\shs{$\mainacc$\\$\robustacc$} &\shs{0.95\\0.41}&\shp{0.95\\0.45}&\shp{0.95\\0.42}&\shp{0.95\\0.42}&\shp{0.94\\0.42}&\shp{0.94\\0.45}&\shp{0.94\\0.43}&\shp{0.93\\0.45}&\shp{0.81\\0.43}&\shp{0.66\\0.51}&\shs{0.67\\0.49}&\shs{0.60\\0.17}&\shs{0.60\\0.13}&\shp{0.59\\0.21}&\shs{0.57\\0.24}&\shs{0.55\\0.27}&\shs{0.53\\0.29}&\shs{0.50\\0.36}&\shs{0.46\\0.40}&\shs{0.39\\0.37}&\shs{0.36\\0.43}&\shs{0.34\\0.43}\\\addlinespace
& \shs{18\\80} &\shs{$\mainacc$\\$\robustacc$} &\shs{0.95\\0.41}&\shp{0.95\\0.49}&\shp{0.95\\0.47}&\shp{0.95\\0.48}&\shp{0.94\\0.49}&\shp{0.94\\0.46}&\shp{0.93\\0.46}&\shs{0.92\\0.37}&\shp{0.72\\0.48}&\shs{0.39\\0.39}&\shs{0.38\\0.41}&\shs{0.60\\0.17}&\shs{0.59\\0.15}&\shs{0.57\\0.16}&\shs{0.55\\0.23}&\shs{0.52\\0.28}&\shs{0.50\\0.34}&\shs{0.46\\0.28}&\shs{0.40\\0.29}&\shs{0.30\\0.35}&\shs{0.25\\0.33}&\shs{0.22\\0.31}\\
\midrule\parbox[t]{2mm}{\multirow{7}{*}{\rotatebox[origin=c]{90}{SurFree with JPEG}}}
& \shs{85} &\shs{$\mainacc$\\$\robustacc$} &\shs{0.95\\0.21}&\shp{0.95\\0.26}&\shp{0.95\\0.33}&\shp{0.95\\0.52}&\shp{0.95\\0.36}&\shp{0.94\\0.40}&\shp{0.94\\0.67}&\shp{0.94\\0.44}&\shp{0.92\\0.83}&\shs{0.92\\0.65}&\shs{0.92\\0.73}&\shs{0.60\\0.52}&\shs{0.60\\0.33}&\shs{0.59\\0.43}&\shp{0.59\\0.79}&\shs{0.58\\0.47}&\shp{0.58\\0.90}&\shs{0.57\\0.51}&\shp{0.56\\0.94}&\shs{0.56\\0.85}&\shs{0.56\\0.49}&\shs{0.56\\0.85}\\\addlinespace
& \shs{75} &\shs{$\mainacc$\\$\robustacc$} &\shs{0.95\\0.21}&\shs{0.95\\0.21}&\shp{0.95\\0.31}&\shp{0.95\\0.37}&\shp{0.94\\0.57}&\shp{0.94\\0.49}&\shp{0.94\\0.40}&\shp{0.93\\0.61}&\shp{0.91\\0.84}&\shs{0.90\\0.71}&\shs{0.90\\0.67}&\shs{0.60\\0.52}&\shs{0.60\\0.36}&\shs{0.59\\0.43}&\shs{0.59\\0.45}&\shp{0.58\\0.78}&\shp{0.57\\0.81}&\shs{0.56\\0.50}&\shs{0.56\\0.51}&\shs{0.55\\0.60}&\shp{0.54\\0.89}&\shs{0.54\\0.82}\\\addlinespace
& \shs{60} &\shs{$\mainacc$\\$\robustacc$} &\shs{0.95\\0.21}&\shp{0.95\\0.42}&\shp{0.95\\0.23}&\shp{0.95\\0.41}&\shp{0.94\\0.50}&\shs{0.94\\0.31}&\shp{0.94\\0.44}&\shs{0.93\\0.36}&\shp{0.90\\0.77}&\shs{0.87\\0.60}&\shs{0.87\\0.60}&\shs{0.60\\0.52}&\shp{0.60\\0.65}&\shs{0.59\\0.53}&\shs{0.58\\0.65}&\shs{0.57\\0.52}&\shp{0.56\\0.89}&\shs{0.55\\0.60}&\shs{0.53\\0.50}&\shs{0.52\\0.50}&\shs{0.52\\0.55}&\shs{0.52\\0.83}\\\addlinespace
& \shs{50} &\shs{$\mainacc$\\$\robustacc$} &\shs{0.95\\0.21}&\shs{0.95\\0.19}&\shp{0.95\\0.30}&\shp{0.95\\0.55}&\shp{0.94\\0.46}&\shs{0.94\\0.32}&\shs{0.94\\0.38}&\shs{0.93\\0.55}&\shp{0.88\\0.75}&\shs{0.85\\0.60}&\shs{0.85\\0.64}&\shs{0.60\\0.52}&\shp{0.60\\0.73}&\shp{0.59\\0.80}&\shs{0.58\\0.49}&\shp{0.57\\0.82}&\shs{0.56\\0.56}&\shp{0.54\\0.88}&\shs{0.52\\0.52}&\shs{0.51\\0.56}&\shs{0.50\\0.60}&\shs{0.50\\0.83}\\
\bottomrule
        \end{tabular}
    
    \caption{Selected results for $\robustacc$ and $\mainacc$ based on the defense parameter $\nu$ and the threshold $\tau$ in the CIFAR-10 and the CIFAR-100 datasets. 
    Improvements beyond the Pareto frontier achieved in vanilla constructs (i.e., when $\tau=1.0$) are {bolded}.} 
    \label{tab:results}
\end{table*}

\begin{table}[t]
    \centering
    \footnotesize
    \setlength{\tabcolsep}{5.5pt}
        \begin{tabular}{cc|c|ccHccHHcHHc}
            \toprule
             & Dataset        &                 & \multicolumn{11}{c}{ImageNet}                                                                                                                                                                                       \\
             & $\nu$ / $\tau$ &                 & 0.0                            & 0.3                             & 0.4                           & 0.5 & 0.6 & 0.7 & 0.8 & 0.9 & 0.97 & 0.99 & 1.0 \\
            \midrule
\parbox[t]{2mm}{\multirow{5}{*}{\rotatebox[origin=c]{90}{\shortstack{PSJA\\with RND}}}}
& \shs{0.01} &\shs{$\mainacc$\\$\robustacc$} &\shs{0.81\\0.97}&\shs{0.81\\0.97}&\shs{0.81\\0.97}&\shp{0.80\\0.98}&\shs{0.80\\0.97}&\shs{0.80\\0.97}&\shs{0.80\\0.97}&\shp{0.80\\0.98}&\shs{0.80\\0.97}&\shs{0.80\\0.97}&\shs{0.80\\0.97}\\\addlinespace
& \shs{0.02} &\shs{$\mainacc$\\$\robustacc$} &\shs{0.81\\0.97}&\shs{0.81\\0.97}&\shs{0.80\\0.96}&\shp{0.80\\0.98}&\shp{0.80\\0.98}&\shs{0.80\\0.97}&\shs{0.80\\0.97}&\shs{0.80\\0.97}&\shs{0.80\\0.97}&\shs{0.80\\0.97}&\shs{0.80\\0.97}\\\addlinespace
& \shs{0.05} &\shs{$\mainacc$\\$\robustacc$} &\shs{0.81\\0.97}&\shs{0.80\\0.96}&\shs{0.80\\0.94}&\shs{0.79\\0.94}&\shs{0.78\\0.94}&\shs{0.77\\0.90}&\shs{0.77\\0.91}&\shs{0.77\\0.91}&\shs{0.76\\0.92}&\shs{0.76\\0.91}&\shs{0.76\\0.91}\\
\midrule\parbox[t]{2mm}{\multirow{5}{*}{\rotatebox[origin=c]{90}{\shortstack{SurFree\\with JPEG}}}}
& \shs{85} &\shs{$\mainacc$\\$\robustacc$} &\shs{0.81\\0.61}&\shp{0.80\\0.77}&\shp{0.80\\0.80}&\shp{0.79\\0.79}&\shp{0.79\\0.79}&\shp{0.79\\0.77}&\shs{0.78\\0.76}&\shs{0.78\\0.78}&\shs{0.78\\0.78}&\shs{0.78\\0.78}&\shs{0.78\\0.78}\\\addlinespace
& \shs{75} &\shs{$\mainacc$\\$\robustacc$} &\shs{0.81\\0.61}&\shp{0.80\\0.82}&\shp{0.80\\0.79}&\shp{0.79\\0.79}&\shs{0.78\\0.76}&\shs{0.77\\0.75}&\shs{0.77\\0.73}&\shs{0.76\\0.70}&\shs{0.76\\0.72}&\shs{0.76\\0.72}&\shs{0.76\\0.72}\\\addlinespace
& \shs{60} &\shs{$\mainacc$\\$\robustacc$} &\shs{0.81\\0.61}&\shp{0.80\\0.80}&\shp{0.79\\0.79}&\shp{0.79\\0.76}&\shs{0.78\\0.76}&\shs{0.76\\0.74}&\shs{0.75\\0.68}&\shs{0.74\\0.65}&\shs{0.74\\0.64}&\shs{0.74\\0.64}&\shs{0.74\\0.64}\\
\bottomrule
        \end{tabular}
    
    \caption{Selected results for $\robustacc$ and $\mainacc$ based on the defense parameter $\nu$ and the threshold $\tau$ in the ImageNet dataset. 
  Our complete results are included in the Appendix.}
    \label{tab:results_2}
\end{table}

\paragraph{Models.}
{
    For CIFAR-10, we use a DenseNet-121 model with an accuracy of $95\%$; and a ResNet-50 for CIFAR-100 and ImageNet with an accuracy of $60\%$ and $81\%$, respectively.
    All our models are calibrated~\cite{guoCalibrationModernNeural}.
}

\subsection{Metrics}
\label{sec:metrics}
\paragraph{Attack success rate and robust accuracy.}
Let~$S\subset \setX \times \setY$ denote the set of (labeled) genuine samples provided to the attacker, let~$n := \size{S}$,
let~$Q$ denote the query budget,
and let~$\varepsilon$ denote the distortion budget.
To compute the attack success rate ($\mathsf{ASR}$) in practice. we determine the number of successful adversarial samples generated by the attacker:
\begin{equation}
    n_{\mathrm{succ}} := \size{ \{ x \in \adv(S)\ |~\adv_{x_0}(x)>0 \land \dist{x-x_0}_p \leq \varepsilon \}},
\end{equation}
where~$\adv(S)$ denotes the set of candidate adversarial samples output by~$\adv$ in a run of the attack on input~$S$.
The $\mathsf{ASR}$ is defined as $\advsuccrate := \nicefrac{n_{\mathrm{succ}}}{n}$, i.e., the ratio of successful adversarial samples.
The complement of the $\advsuccrate$ is the robust accuracy ($\robustacc$) of the classifier, i.e., $\robustacc = 1 - \advsuccrate$.

\paragraph{CA-RA Pareto frontier.}
In multi-objective optimization problems, there is often no solution that maximizes all objective functions simultaneously.
The Pareto frontier emerges as an effective tool to evaluate tradeoffs between the various objectives.
To evaluate the accuracy-robustness tradeoffs of different defenses, we consider $\mainacc$ and $\robustacc$ as our objectives and empirically determine the Pareto frontier.
Specifically, we consider the following optimization problem:
\begin{equation}
    \max_{\omega \in \Omega} (\mainacc(\omega),\robustacc(\omega)),
\end{equation}
where $\mainacc(\omega)$ and~$\robustacc(\omega)$ denote the clean accuracy and robust accuracy of a given (protected) classifier as functions of the defense parameter~$\omega$ -- for a fixed attack and experiment setting.

A solution~$\omega^*$
is \emph{Pareto optimal} if there exists no other solution that improves all objectives simultaneously.
Formally, given two solutions~$\omega_1$ and~$\omega_2$, we write~$\omega_1 \succ \omega_2$ if~$\omega_1$ dominates~$\omega_2$, i.e., if
$\mainacc(\omega_1) \geq \mainacc(\omega_2) \land \robustacc(\omega_1) \geq \robustacc(\omega_2)$.
The \emph{Pareto frontier} is the set of Pareto-optimal solutions:
\begin{equation}
    PF(\Omega) = \{ \omega^*\in \Omega \ | \ \nexists \omega \in \Omega\text{ s.t. } \omega\succ \omega^*\}.
\end{equation}

\subsection{Evaluation Setup} \label{sec:eval_setup}
To evaluate the effectiveness of our approach, we instantiate our proposal with three existing defenses~$\defense$, namely RND, RCR, and JPEG compression, and empirically measure the accuracy and robust accuracy of the resulting classifiers~$F_{\defense,\tau}$ against state-of-the-art decision-based attacks PSJA and SurFree in the untargeted setting.

We generate adversarial examples under the~$l_2$ norm constraint
    {with maximum distortion~$\varepsilon = 3$ for both CIFAR-10 and CIFAR-100, assuming a query budget of ${Q=20,000}$}. {Due to the significantly larger input dimension of ImageNet,}
{we allow for a higher query budget of ${Q=40,000}$ in accordance with previous works and we select a comparable $\varepsilon = 21$.}
We evaluate our proposal with ${\tau \in \{0.0, 0.3, 0.4, 0.5, 0.6, 0.7, 0.8, 0.9, 0.97, 0.99, 1.0\}}$. In extreme cases, $\tau = 0$ means that the defense is never activated, while $\tau = 1$ triggers the defense on all inputs.

In combination with the aforementioned thresholds, we select a spectrum of noise levels to achieve a diverse set of $\mainacc$ and $\robustacc$. More concretely, we select the noise parameter $\nu$ of the RND defense from the set $\{0.01, 0.02, 0.05, 0.07, 0.08, 0.1\}$,  for RCR ${\nu\in\{29, 27, 26, 25, 22, 18, 14\}}$ for CIFAR-10/100 and ${\nu\in\{200, 176, 152, 128, 104, 80, 56\}}$ for ImageNet. Lastly, we use a quality setting of $\nu \in \{85, 75, 60, 50, 35, 25, 10\}$ for the JPEG defense.

Throughout our evaluation, we evaluate PSJA against randomized defenses, i.e., RND, RCR, and SurFree against the deterministic defense, i.e., JPEG.
We always measure the clean accuracy (or $\mainacc$) of the classifier over the entire test set of each respective dataset and the robust accuracy ($\robustacc$) on $n=100$ randomly selected images from the test set that are correctly classified by the undefended classifier.

Our experiments have been conducted on a server equipped with two AMD EPYC 7542 CPUs, two Nvidia A40 GPUs, and 256GB RAM. The system runs Ubuntu 22.04., Python 3.9, PyTorch 1.13.0, and CUDA 11.7.

\subsection{Evaluation Results} \label{sec:eval}

\begin{figure*}[tb]
    \centering
        \begin{subfigure}[t]{.24\textwidth}
        \centering
        \includegraphics[width=\linewidth]{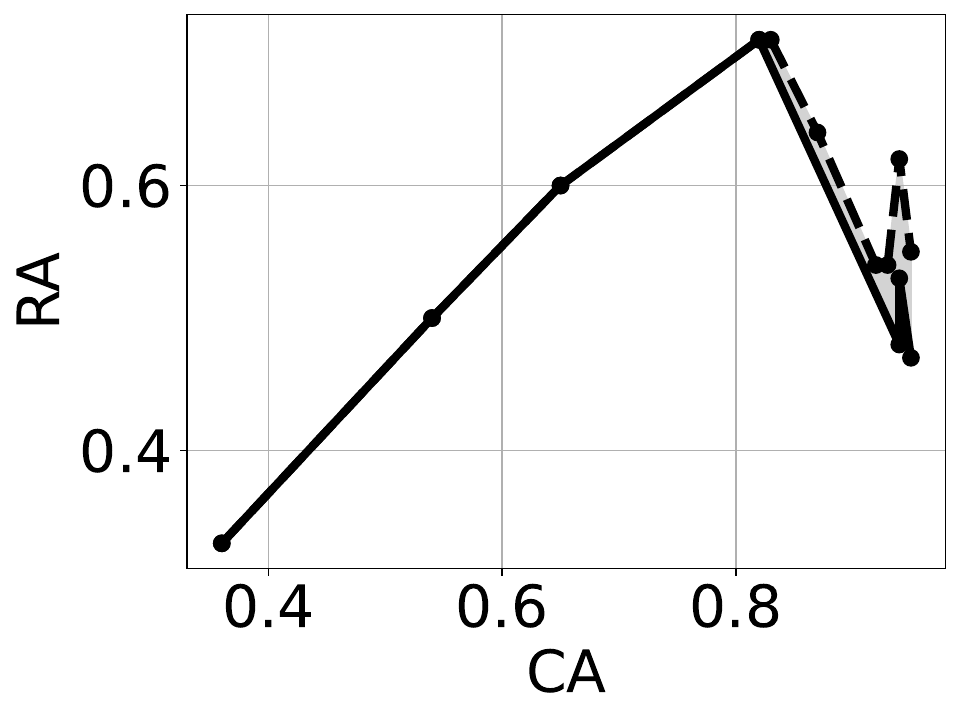}
        \caption{PSJA, CIFAR-10, $\defense$=RND}
   	    \label{fig:ct-cifar10-psja-rnd}
    \end{subfigure}
        \begin{subfigure}[t]{.24\textwidth}
        \centering
        \includegraphics[width=\linewidth]{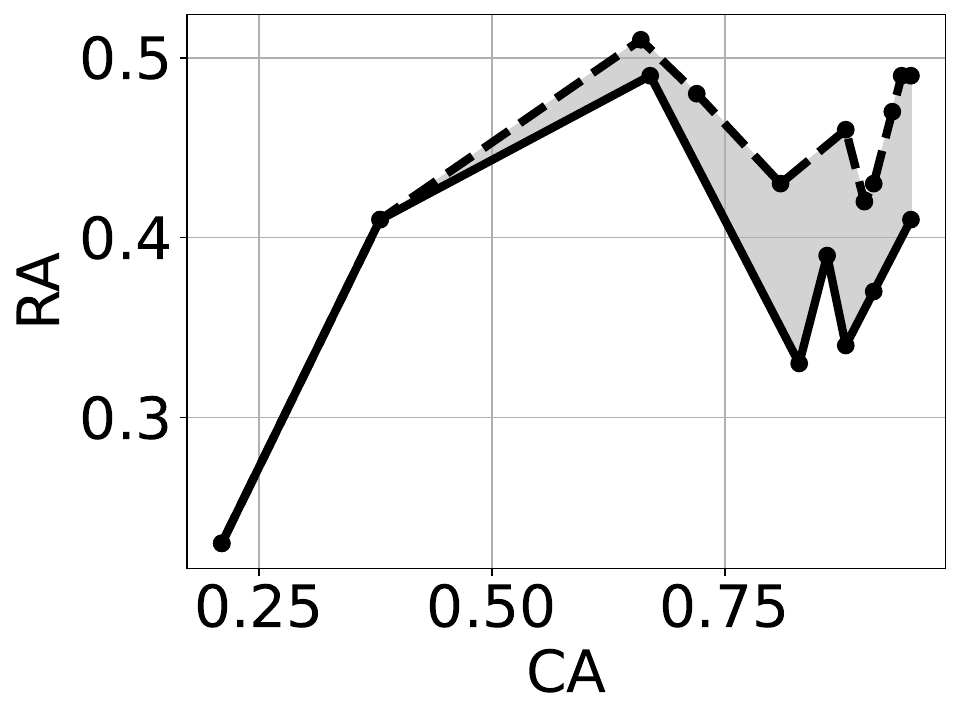}
        \caption{PSJA, CIFAR-10, $\defense$=RCR}
        \label{fig:ct-cifar10-psja-rcr}
    \end{subfigure}
        \begin{subfigure}[t]{.24\textwidth}
        \centering
        \includegraphics[width=\linewidth]{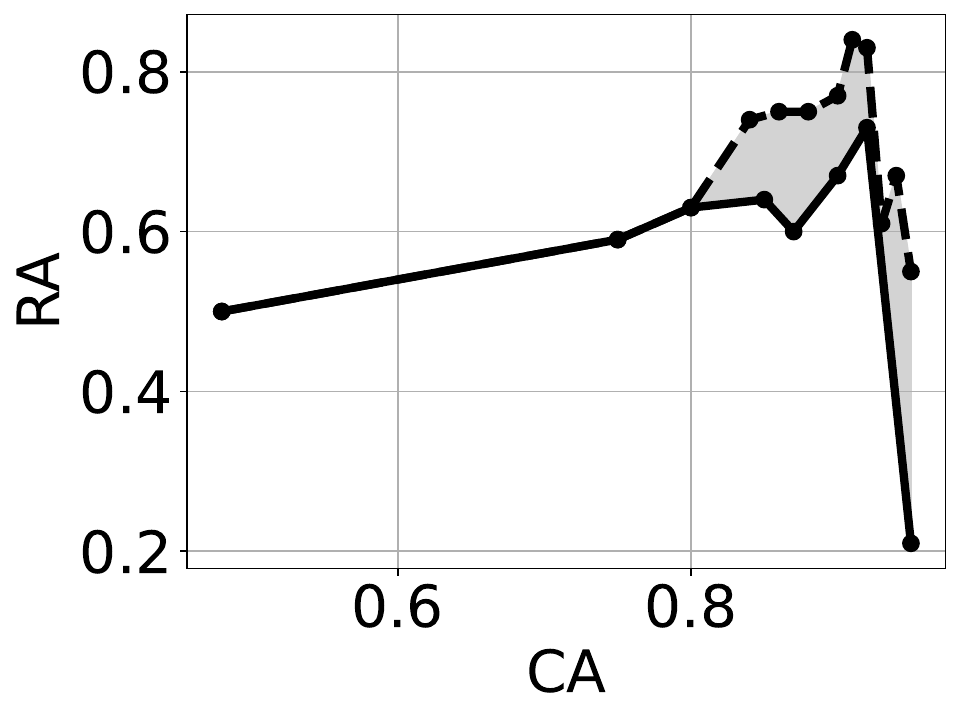}
        \caption{SurFree, CIFAR-10, $\defense$=JPEG}
        \label{fig:ct-cifar10-surfree-jpeg}
    \end{subfigure}
         \begin{subfigure}[t]{.24\textwidth}
        \centering
        \includegraphics[width=\textwidth]{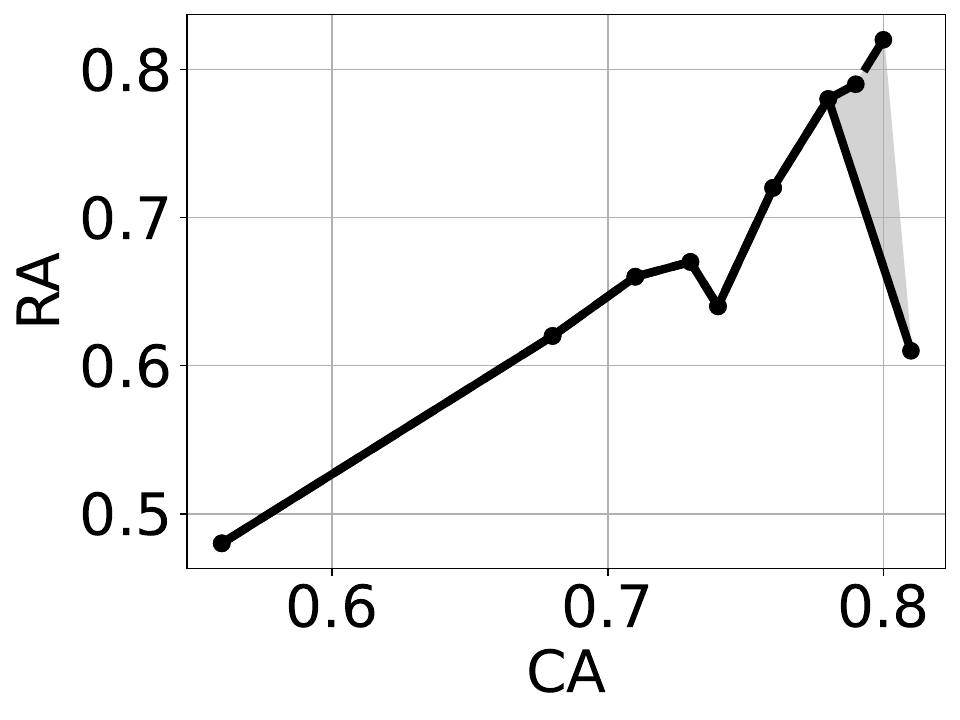}
        \caption{SurFree, ImageNet, $\defense$=JPEG}
        \label{fig:ct-imagenet-surfree-jpeg}
    \end{subfigure}
        \begin{subfigure}[t]{.24\textwidth}
        \centering
        \includegraphics[width=\linewidth]{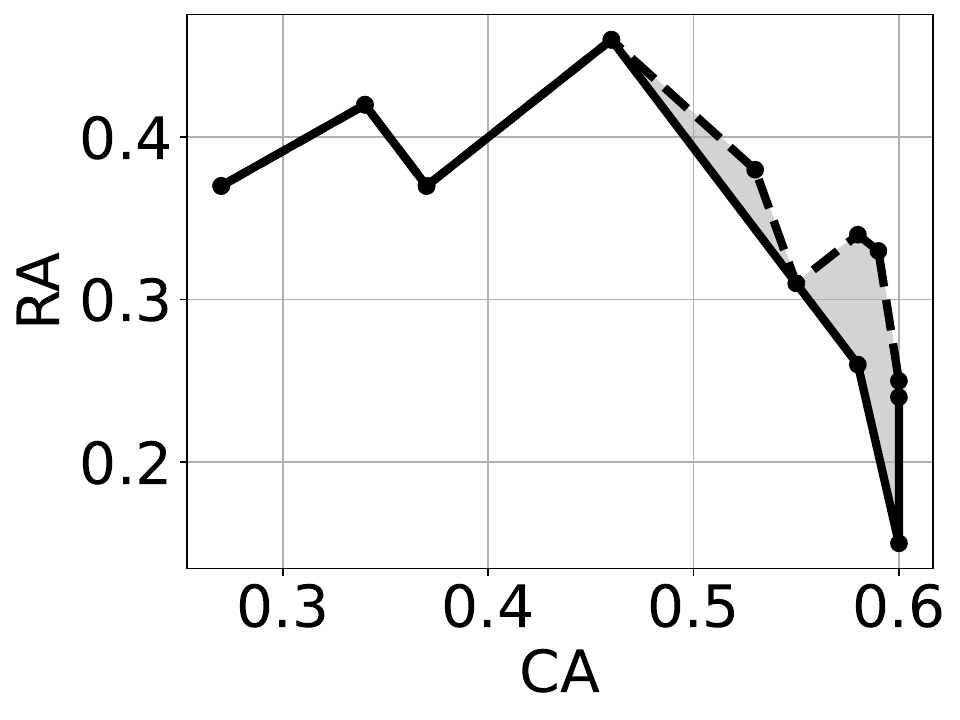}
        \caption{PSJA, CIFAR-100, $\defense$=RND}
        \label{fig:ct-cifar100-psja-rnd}
    \end{subfigure}
        \begin{subfigure}[t]{.24\textwidth}
        \centering
        \includegraphics[width=\linewidth]{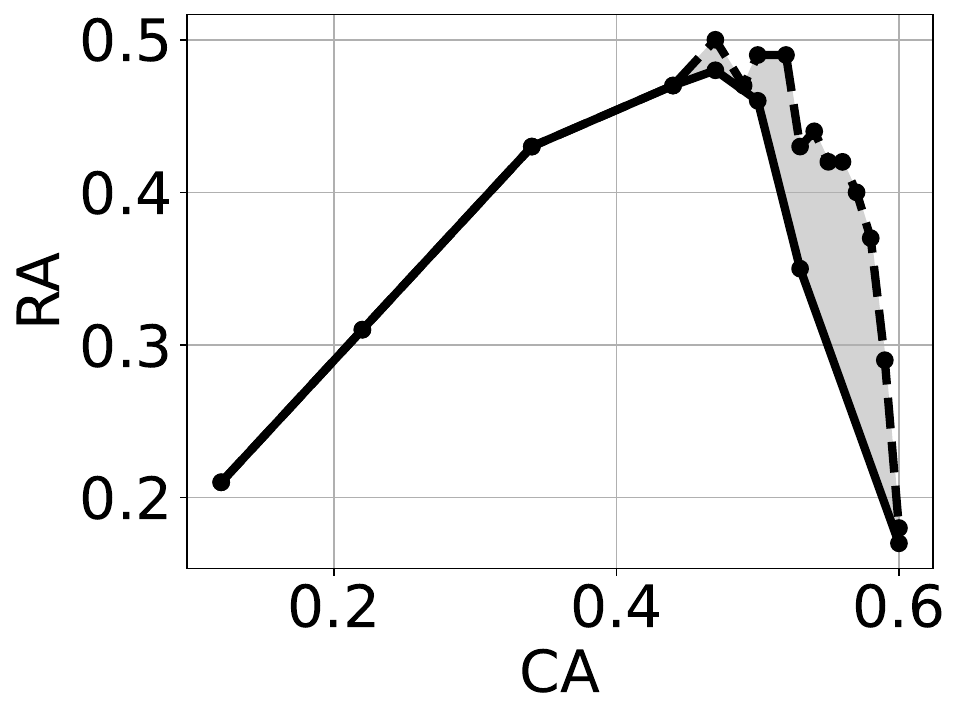}
        \caption{PSJA, CIFAR-100, $\defense$=RCR}
        \label{fig:ct-cifar100-psja-rcr}
    \end{subfigure}
        \begin{subfigure}[t]{.24\textwidth}
        \centering
        \includegraphics[width=\linewidth]{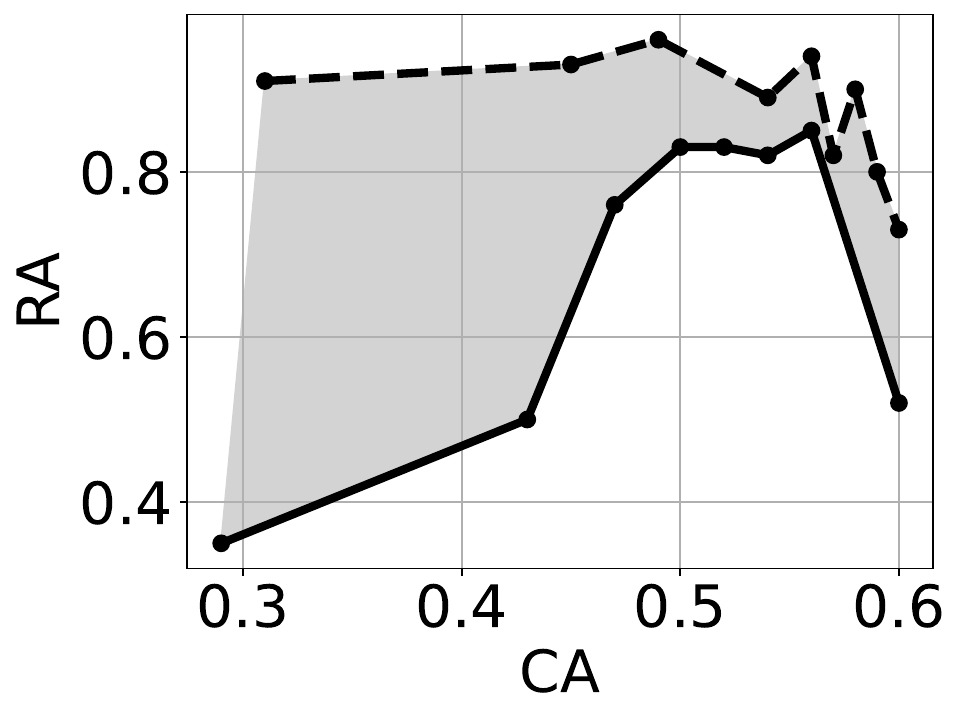}
        \caption{SurFree,CIFAR-100,$\defense$=JPEG}
        \label{fig:ct-cifar100-surfree-jpeg}
    \end{subfigure}
    \begin{subfigure}[t]{.24\textwidth}
        \centering
        \includegraphics[width=\linewidth]{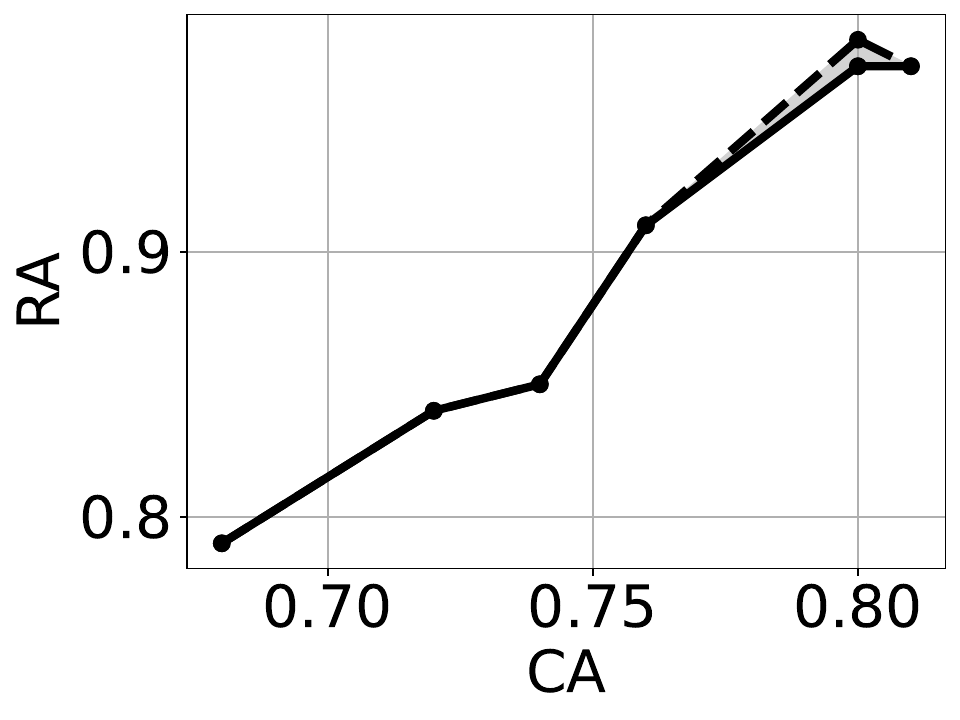}
        \caption{PSJA, ImageNet, $\defense$=RND}
        \label{fig:ct-imagenet-psja-rnd}
    \end{subfigure}\\
    \caption{Pareto frontier when applying our approach in conjunction with RND, RCR, and JPEG on CIFAR-10, CIFAR-100, and ImageNet using the attacks PSJA and SurFree.}
    \label{fig:ct_pareto_frontier}
\end{figure*}

\paragraph{Accuracy-robustness tradeoffs.}

In \Cref{fig:ct_pareto_frontier}, we use the Pareto frontier to showcase the tradeoff improvements of our proposal compared to the baseline relying only on the defense parameter~$\nu$.
    {We plot the points~$(\mainacc(\omega^*),\robustacc(\omega^*))$ for empirically-determined Pareto-optimal solutions~$\omega^*$ for the baseline defense ($\omega^* = \nu^*$, solid line) and for our approach ($\omega^* = (\nu^*,\tau^*)$, dotted line). The accuracy-robustness tradeoffs achieved by our proposal for selected threshold values are highlighted in grey.  The specific combinations of $\nu, \tau$ that outperform the baseline tradeoffs (${\tau=1.0}$) are highlighted in Table~\ref{tab:results} and \Cref{tab:results_2}. Note that for RCR the crop size for CIFAR-10/100 and ImageNet are shown due to the differently sized input dimensions.}

\paragraph{Tradeoffs in PSJA.}

When evaluated against PSJA on the CIFAR-10 dataset, our proposal combined with the RND defense can provide an improvement in $\robustacc$ while preserving the ~$\mainacc$ of the baseline. For example, it improves~$\robustacc$ by $8\%$ (out of a maximum of $9\%$)
for $\mainacc=0.95$ and $\robustacc$ by up to~$9\%$ for a tradeoff of at most~$1\%$ decrease in~$\mainacc$. Compared to the baseline ($\tau=1.0,\nu=0.07$), we improve $\mainacc$ by $29\%$, while decreasing $\robustacc$ by just $1\%$ (cf. \Cref{fig:ct-cifar10-psja-rnd}).

When evaluated with the RCR defense, we can observe several improvements in \Cref{fig:ct-cifar10-psja-rcr}. For example, given a $\robustacc$ of $49\%$, our approach can increase $\mainacc$ from $67\%$ to $94\%$.
At a $\robustacc$ of $41\%$, our proposal achieves an $8\%$ improvement in robustness compared to the baseline without compromising $\mainacc$. This is in line with the results for RND.

Our results on CIFAR-100 shown in \Cref{fig:ct-cifar100-psja-rnd,fig:ct-cifar100-psja-rcr} are consistent with our CIFAR-10 results. More precisely, we measure an improvement of $\robustacc$ of up to $8\%$ for a $\mainacc$ of $58\%$ when evaluating with the RND defense. Compared to the baseline with a $\mainacc / \robustacc$ of $37\%$, our approach improves the $\mainacc$ by $16\%$ and $\robustacc$ by $1\%$.
When considering the RCR defense in \Cref{fig:ct-cifar100-psja-rcr}, our approach
can improve $\mainacc$ by $19\%$ for the same $\robustacc$ of $43\%$. When  $\mainacc=0.53$, our approach yields an improvement in $\robustacc$ by $8\%$.
For ImageNet, we see an improvement of up to $1\%$ in $\robustacc$ (cf. \Cref{fig:ct-imagenet-psja-rnd})---we discuss these results in more detail in the Appendix.

\paragraph{Tradeoffs in SurFree.}
\Cref{fig:ct-cifar10-surfree-jpeg}, \Cref{fig:ct-cifar100-surfree-jpeg}, and \Cref{fig:ct-imagenet-surfree-jpeg} depict the evaluation results against the SurFree attack, for which our proposal can obtain consistent improvements compared to the baseline. At no decrease in $\mainacc$, we observe significant improvements in $\robustacc$. For CIFAR-10, it increases by $34\%$ and for CIFAR-100 by $21\%$. This trend is also observable for ImageNet---we obtain an increase of $\robustacc$ of $21\%$ with a negligible decrease of $\mainacc$ by $1\%$.
When fixing $\robustacc$ to $67\%$, we obtain an improvement in $\mainacc$ of $4\%$ for CIFAR-10. Moreover, we see an improvement of $\mainacc$ by $3\%$ at a fixed $\robustacc=0.82$ for CIFAR-100.

\paragraph{Comparison to related work.}
We now compare our proposal to various training-time defenses, which require the costly retraining of a model and cannot be retrofitted to existing models. We compare against OAT~\cite{wangOnceforallAdversarialTraining2020} because it is the only reconfigurable, i.e., adjustment of accuracy-robustness tradeoff after training, adversarial-training solution with an open-source implementation. In addition, we include traditional, non-adjustable, training-time defenses, such as PNI~\cite{he2019parametric}, RSE~\cite{liu2018towards}, and a state-of-the-art robust model (AT)~\cite{gowal2020uncovering} from RobustBench for further comparison. We also include results from the previously considered vanilla inference-time defenses RCR and JPEG with $\tau=1.0$, i.e., traditional deployment of defenses for this comparison.

\begin{table}[tb]
    \footnotesize
    \centering
        \begin{tabular}{c|c|cccc|cc}
            \toprule
             &            & \multicolumn{4}{c|}{Training-time} & \multicolumn{2}{c}{Inference-time}                                                           \\\midrule
             &   $\sim$       &                                   &                               &        &                 & Baseline     &            \\
             & $\mainacc$ & OAT                               & PNI                           & RSE    & AT              & ($\tau=1.0$) &  Ours      \\
            \midrule
            \parbox[t]{2mm}{\multirow{3}{*}{\rotatebox[origin=c]{90}{PSJA}}}
             & 0.91       & 0.67                              & --                            & {0.20} & {\textbf{0.83}} & {0.37}       & {0.43}        \\
             & 0.88       & \textbf{0.77}                     & --                            & --     & --              & {0.34}       & {0.46}        \\
             & {0.84}     & --                                & \textbf{0.98}                 & --     & --              & {0.33}       & {0.43}        \\
            \midrule
            \parbox[t]{2mm}{\multirow{3}{*}{\rotatebox[origin=c]{90}{SurFree}}}
             & 0.91       & 0.47                              & --                            & {0.19} & {0.76}          & {0.73}       & \textbf{0.84} \\
             & 0.88       & \textbf{0.82}                     & --                            & --     & --              & {0.60}       & {0.75}        \\
             & 0.84       & --                                & \textbf{0.97}                 & --     & --              & {0.64}       & {0.74}        \\

            \bottomrule
        \end{tabular}
    \caption{$\robustacc$ for state-of-the-art defenses on CIFAR-10. The baseline instantiates PSJA with $\defense$ = RCR and SurFree with $\defense$ = JPEG.}
    \label{tab:results_training_based}
\end{table}

In \Cref{tab:results_training_based}, we evaluate $\robustacc$ at those values of $\mainacc$ obtained with these defenses against PSJA and SurFree on CIFAR-10. Note that some values of $\mainacc$ cannot be achieved with some defenses. In these cases, we report the $\robustacc$ that results in the closest (by up to $1-2\%$) $\mainacc$; when such a close estimate cannot be found, we do not report any value for $\robustacc$. 

Against SurFree, our proposal outperforms all training-/inference-time-based defenses, outperforming AT by $8\%$, while being completely training-free. For $\mainacc=0.88$, we observe a difference to the $\robustacc$ of OAT of just $7\%$. For ${\mainacc=0.84}$, a larger difference of $23\%$ in $\robustacc$ to the one of PNI can be measured. We conclude that, although it is training-free, our proposal manages to outperform or closely perform similarly to existing defenses against SurFree. In the case of PSJA, training-time defenses outperform our approach by a minimum of $31\%$ in $\robustacc$. Compared to OAT, this difference is reduced to $24\%$. Our approach, however, results in significant improvements compared to all training-free defenses.

\subsection{Ablation Study}

Our proposal relies on two main parameters: the confidence threshold~$\tau$ and the genuine defense's parameter $\nu$.
To evaluate the impact of each of these parameters on robustness, we present an ablation study.
We detail our results in~\Cref{tab:results} and \Cref{tab:results_2}, where we show the achieved $\mainacc$ (top of each row) and $\robustacc$ (bottom of each row) for different values of~$\tau$ and~$\nu$.
We highlight the values that result in an improvement of the Pareto frontier in bold. When $\tau=1.0$, our proposal instantiates the baseline since the defense is always enabled.

\begin{figure}[tb]
    \centering
    \begin{subfigure}[t]{.23\textwidth}
        \centering
        \includegraphics[width=\linewidth]{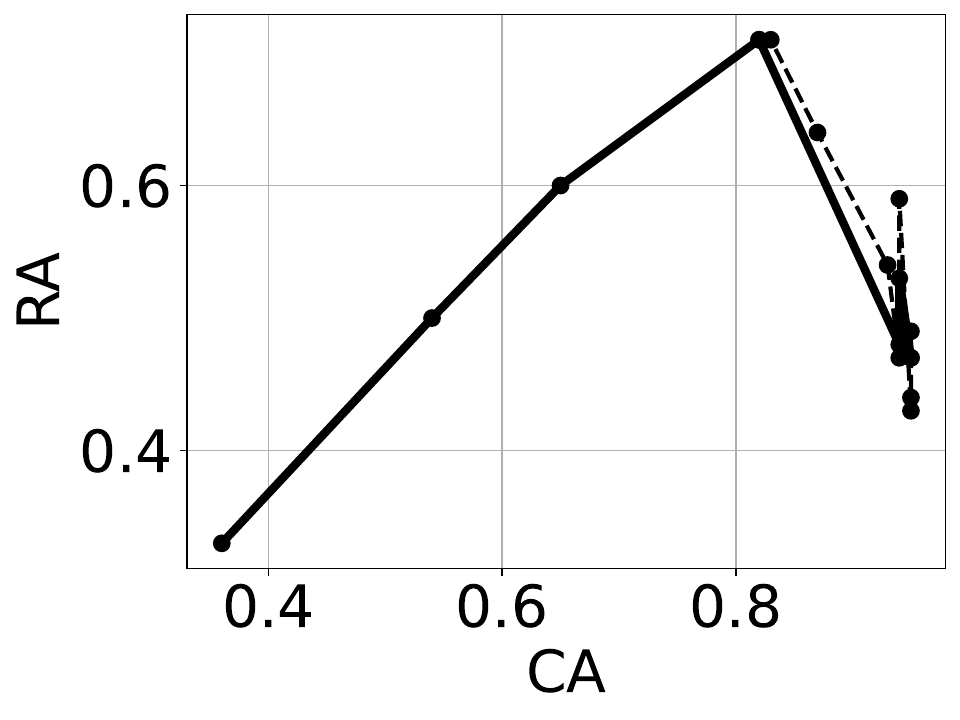}
        \caption{PSJA, $\defense$=RND, $\nu=0.05$}
        \label{fig:ct-cifar10-ablation-psja_rnd_best_noise}
    \end{subfigure}
    \begin{subfigure}[t]{.23\textwidth}
        \centering
        \includegraphics[width=\linewidth]{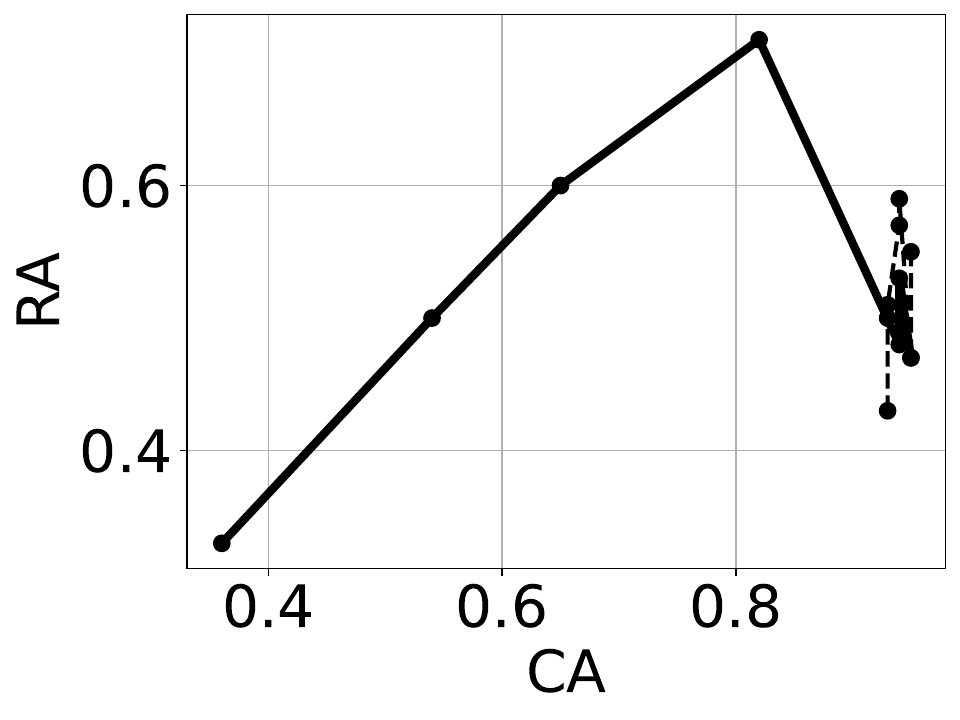}
        \caption{PSJA, $\defense$=RND, $\tau=0.8$}
        \label{fig:ct-cifar10-ablation-psja_rnd}
    \end{subfigure}
    \begin{subfigure}[t]{.23\textwidth}
        \centering
        \includegraphics[width=\linewidth]{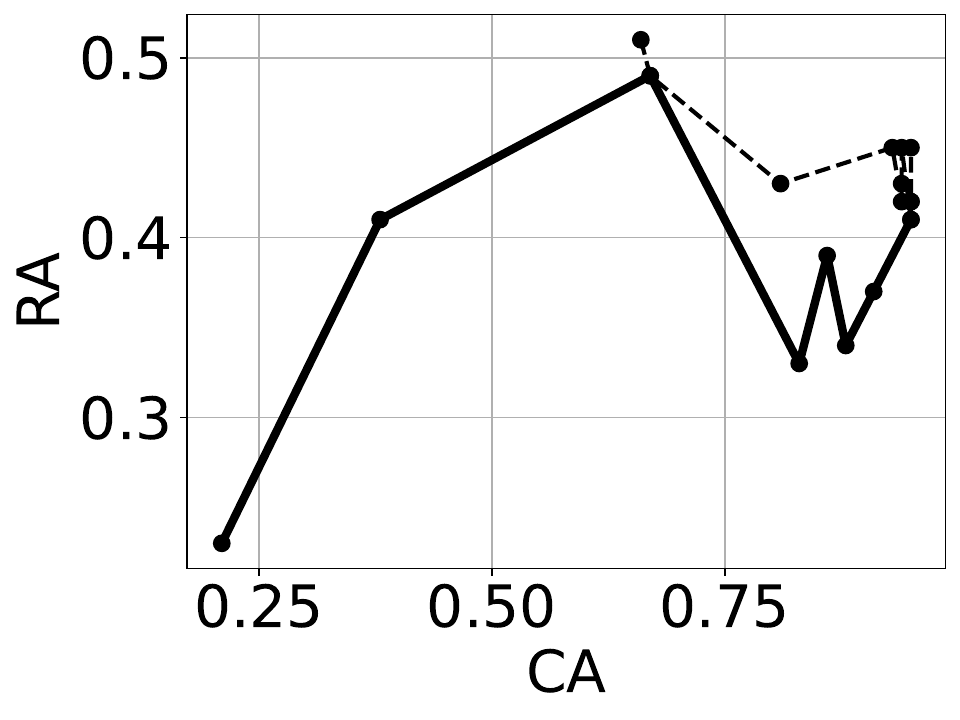}
        \caption{PSJA, $\defense$=RCR, $\nu=22$ }
        \label{fig:ct-cifar10-ablation-psja_rcr_best_noise}
    \end{subfigure}
    \begin{subfigure}[t]{.23\textwidth}
        \centering
        \includegraphics[width=\linewidth]{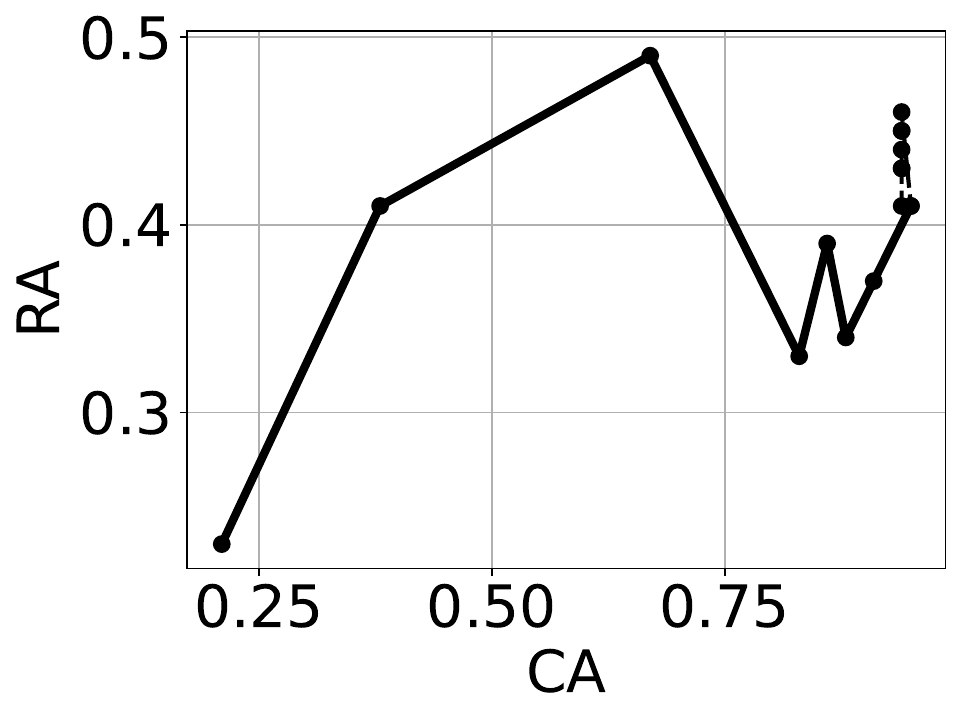}
        \caption{PSJA, $\defense$=RCR, $\tau=0.7$ }
        \label{fig:ct-cifar10-ablation-psja_rcr}
    \end{subfigure}\\
    \begin{subfigure}[t]{.23\textwidth}
        \centering
        \includegraphics[width=\linewidth]{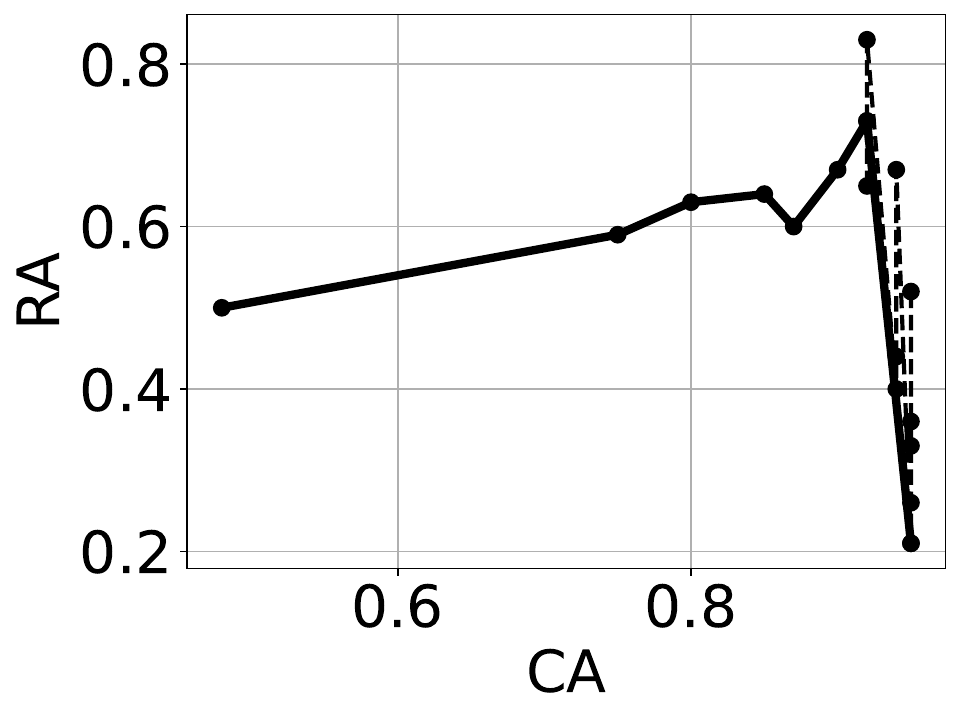}
        \caption{SurFree, $\defense$=JPEG, $\nu=85$ }
        \label{fig:ct-cifar10-ablation-surfree_jpeg_best_noise}
    \end{subfigure}   
    \begin{subfigure}[t]{.23\textwidth}
        \centering
        \includegraphics[width=\linewidth]{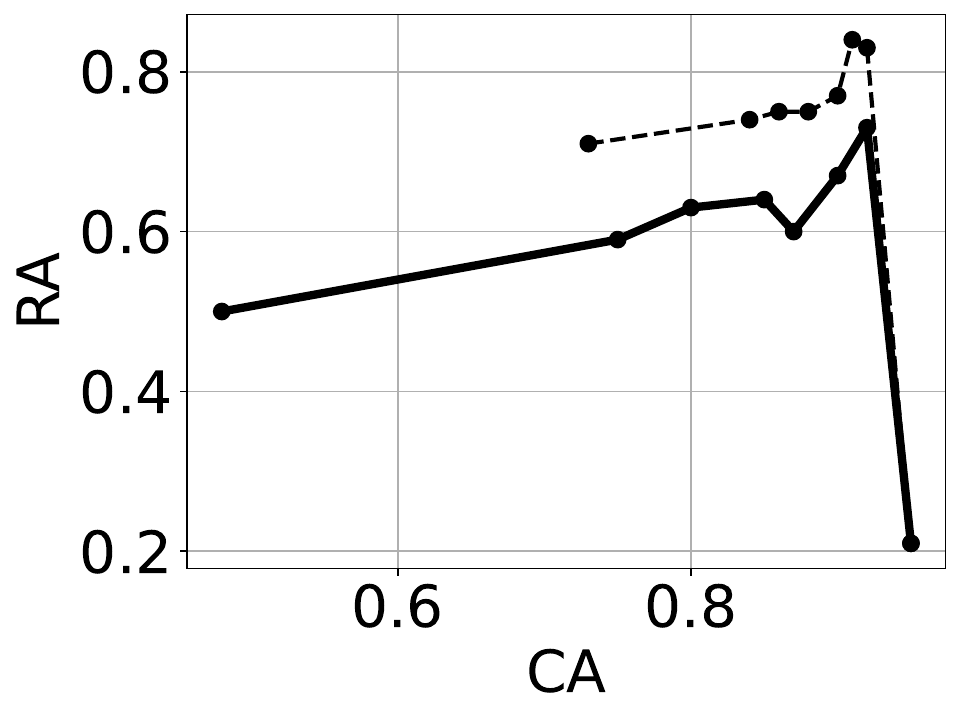}
        \caption{SurFree, $\defense$=JPEG, ${\tau=0.97}$ }
        \label{fig:ct-cifar10-ablation-surfree_jpeg}
    \end{subfigure}
    \caption{Pareto frontier for various $\tau$ and $\nu$ for CIFAR-10.}
    \label{fig:ct_pareto_frontier_ablation_best_tau}
\end{figure}

\paragraph{Impact of ${\tau}$.}
We vary the value of $\tau \in \{0.0, 0.3, 0.4, 0.5, 0.6, 0.7, 0.8, 0.9, 0.97, 0.99, 1.0\}$.
For the CIFAR-10 dataset, we first consider the RND defense with noise $\nu=0.05$, which offers the highest $\robustacc=0.71$ at $\mainacc=0.82$ and the best tradeoffs. As we reduce the value of~$\tau$, we observe a significant improvement in $\mainacc$, increasing from $82\%$ when $\tau \geq 0.99$ to $87\%$ when $\tau=0.97$. By further reducing $\tau$ to $0.8$, our proposal achieves $94\%$ $\mainacc$, fairly close to the original $\mainacc$ of $95\%$.
On the other hand, $\robustacc$ starts at $71\%$ for the baseline, remains consistent at $71\%$ when $\tau=0.99$, and decreases to $64\%$ when $\tau=0.97$. For~$\tau\leq 0.8$, $\robustacc$ remains at $59\%$ and then drops sharply to a $\robustacc$ as low as $43\%$.  The Pareto frontier for the fixed $\nu$ and varying $\tau$ can be found in \Cref{fig:ct-cifar10-ablation-psja_rnd_best_noise}.

For RCR, we show the Pareto frontier for $\nu=22$ in \Cref{fig:ct-cifar10-ablation-psja_rcr_best_noise} and for JPEG for $\nu=85$ in \Cref{fig:ct-cifar10-ablation-surfree_jpeg_best_noise}. For both defenses, we observe consistent improvements in the baseline tradeoffs. We argue that such a noise level introduced by these two defenses is similar to $\nu=0.05$ for the RND defense.
These observations are not exclusive to CIFAR-10 but can also be clearly seen with CIFAR-100 with $\nu=26$ for RCR, while it is observable for $\nu=0.02$ and $\nu=50$ for RND and JPEG, respectively.

\paragraph{Impact of $\nu$.}

We now vary the defense-specific parameter $\nu$ for a fixed $\tau$ for CIFAR-10.
For JPEG we identify a wide range of tradeoffs for $\tau=0.97$, i.e., $\mainacc$ between $73\%$ and $92\%$ and $\robustacc$ between $71\%$ and $83\%$.
In line with previous results, we find an optimal $\tau=0.97$ for which the Pareto frontiers outperform the baseline ($\tau=1.0$) in terms of obtainable tradeoffs, which is highlighted in \Cref{fig:ct-cifar10-ablation-surfree_jpeg}.
For RND, we set $\tau$ to the identified value of $0.8$, as it achieves the best tradeoff across varying $\nu$ (cf. \Cref{fig:ct-cifar10-ablation-psja_rnd}).  It has a negligible impact of $\pm 2\%$ on $\mainacc$ but allows us to control $\robustacc$  between $43\%$ and $59\%$, with the most optimal point found at $\nu=0.05$.
For RCR, we notice a similar behavior at $\tau=0.7$, for which we obtain a $3-5\%$ improvement over the baseline as seen in \Cref{fig:ct-cifar10-ablation-psja_rcr}. For the CIFAR-100, we notice a similar trend for a threshold of $0.7/0.8$.
    {
        For the ImageNet dataset, we consider JPEG with threshold~$\tau=0.3$ and vary the value of $\nu$ as before. When $\nu$ decreases from $85$ to $75$, we observe an improvement in both $\robustacc$ by $5\%$ without hampering $\mainacc$. Reducing~$\nu$ further has a negligible impact on $\mainacc$, while it decreases $\robustacc$ close to the initial value to $77\%$.
    }

    {Overall, by appropriately setting the confidence threshold~$\tau$, our approach can improve the accuracy-robustness tradeoff compared to the baseline for all values of noise~$\nu$.}

\section{Conclusion}

In this paper, we showed that limiting the invocation of an inference-time defense to low-confidence inputs might be sufficient to obstruct the search for adversarial samples in query-based attacks.
Our approach can be applied generically to existing inference-time defenses and is training-free.
We therefore hope to motivate further research in this area.

\section{Acknowledgments}
This work has been co-funded by the Deutsche Forschungsgemeinschaft (DFG, German Research Foundation) under Germany’s Excellence Strategy - EXC 2092 CASA - 390781972, by the German Federal Ministry of Education and Research (BMBF) through the project TRAIN (01IS23027A), and by the European Commission through the HORIZON-JU-SNS-2022 ACROSS project (101097122). Views and opinions expressed are however those of the authors only and do not necessarily reflect those of the European Union. Neither the European Union nor the granting authority can be held responsible for them.

\section{Appendix}

\subsection{Empirical validation of Proposition 1}

To empirically validate that benign samples are typically classified with high confidence and are hence less impacted by our proposal compared to adversarial examples, we consider the distribution of calibrated confidences for the benchmark datasets CIFAR-10, CIFAR-100, and ImageNet.

\begin{figure}[H]
    \center{
        \includegraphics[width=.9\columnwidth]{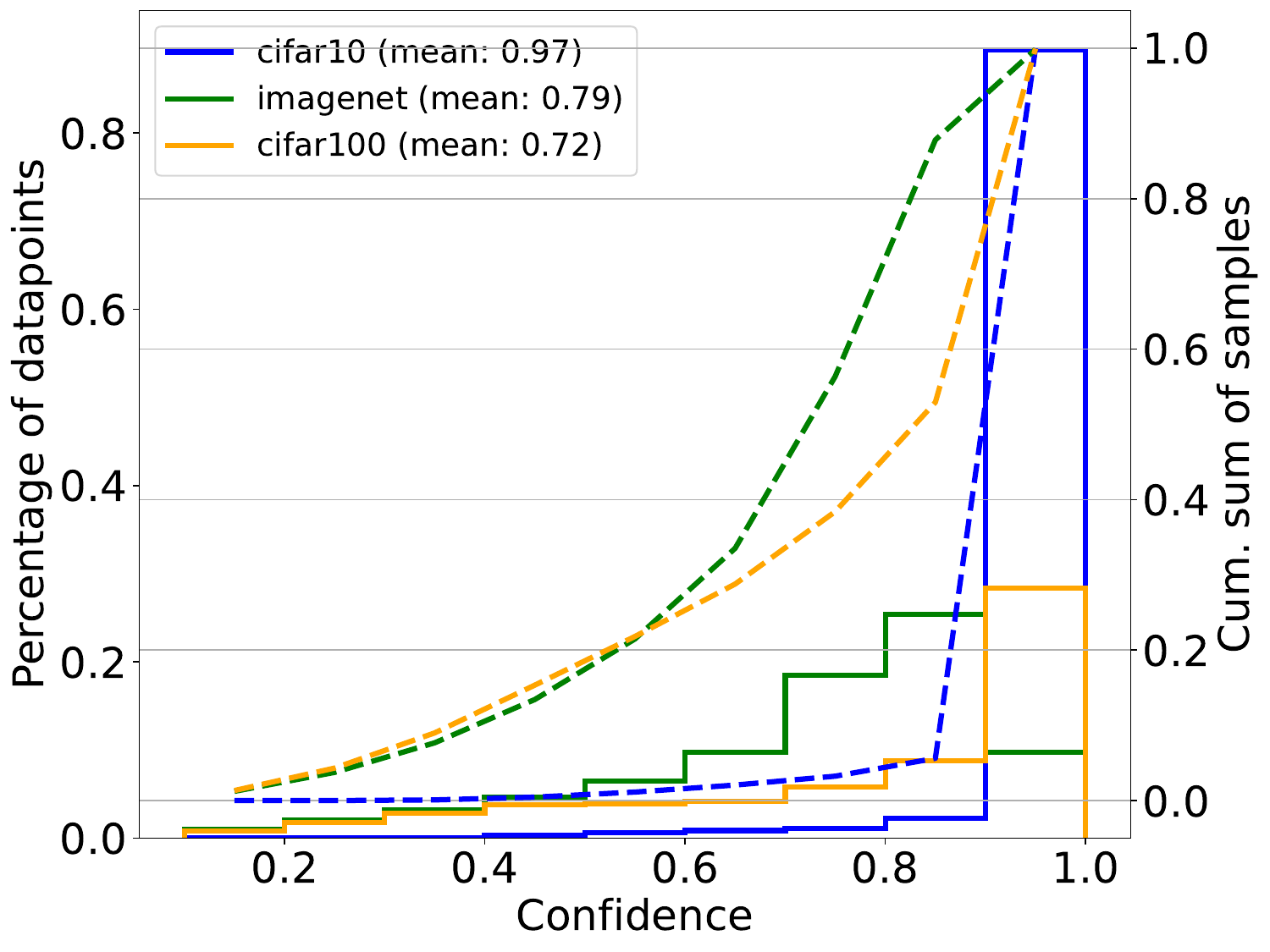}
    }
    \caption{Distribution and cumulative sum of calibrated confidences.}
    \label{fig:validation_hist}
\end{figure}

We show them in \Cref{fig:validation_hist} across the correctly classified data points from the test/validation set (scaled to the size of the set). The plot clearly shows that benign images are generally classified with high confidence. More concretely, we find $90\%$ of images are above $90\%$ in confidence for CIFAR-10 and $80\%$ above $50\%$ in confidence for CIFAR-100 and ImageNet.

\subsection{Proof for Proposition 2}

In the following, we show that low-confidence regions are necessarily explored in the binary-search procedure that is part of a decision-based attack. This is also shown empirically in \Cref{fig:catsdogs}. In line with previous works~\cite{simon-gabrielPopSkipJumpDecisionBasedAttack2021a}, we assume the output probability $p_c^{(t,0)}$ of class $c$ of the interpolation $x$ to have a sigmoidal shape along the line segment $[x_t, x_0]$. This shape is defined as:
\begin{equation}
    p_c^{(t,0)}(k) = \eta + (1-2\eta)\sigma(s(k-z))
\end{equation}
where $\eta$ models an overall noise level, $\sigma(k):=\nicefrac{1}{(1+e^{-4k})}$ is the usual sigmoid, rescaled to get a slope of 1 in its center $z$ when the inverse scale parameter $s$ is equal to 1. 

When performing a binary search between samples of classes $c_t$ and $c_0$, we consider the probabilities $p_{c_t}^{(t,0)}$ and $p_{c_0}^{(t,0)}$. Additionally, we assume that both classes share a common decision boundary.
Given Proposition 1, we have:
\begin{align*}
    p_{c_t}^{(t,0)}(0) \approx p_{c_0}^{(t,0)}(1) \approx  1 \\
    p_{c_t}^{(t,0)}(1) \approx p_{c_0}^{(t,0)}(0) \approx  0
\end{align*}
while the remaining classes have negligible output probabilities $\rho$.
\begin{align}
    \sum_{c=1}^{n}p_c^{(t,0)}(k)                            & = 1                                    \\
    \Leftrightarrow p_{c_t}^{(t,0)}(k) + p_{c_0}^{(t,0)}(k) + \underbrace{\sum_{\substack{c \in [n], \\c \neq c_0 \\\wedge c \neq c_t}}^{}p_c^{(t,0)}(k))}_{\rho} &= 1 \\
    \Leftrightarrow p_{c_t}^{(t,0)}(k) + p_{c_0}^{(t,0)}(k) & = 1-\rho
\end{align}
We now show that $\exists k, \;s.t.\; p_{c_t}^{(t,0)} = p_{c_0}^{(t,0)}$. This would directly signal that a region of low confidence is necessarily explored.

Given the probability function of classes $c_t$ and $c_0$ of a linear interpolation between $[x_t, x_0]$ and interpolation factor $k \in [0,1]$:
\begin{align*}
    p_{c_t}^{(t,0)}(k) & = \eta + (1-2\eta)\sigma(s_0(k-z_0))  \\
    p_{c_0}^{(t,0)}(k) & = \eta + (1-2\eta)\sigma(-s_1(k-z_1))
\end{align*}
with $s_0, s_1 \in \mathbb{R}^{+}, z_0, z_1 \in \mathbb{R}$.
\begin{align*}
    \Rightarrow     p_{c_t}^{(t,0)}(k)                 & = p_{c_0}^{(t,0)}(k)                  \\
    \Leftrightarrow \eta + (1-2\eta)\sigma(s_0(k-z_0)) & = \eta + (1-2\eta)\sigma(-s_1(k-z_1)) \\
    \Leftrightarrow \sigma(s_0(k-z_0))                 & = \sigma(-s_1(k-z_1))                 \\
    \Leftrightarrow \frac{1}{1+e^{-4(s_0(k-z_0))}}     & = \frac{1}{1+e^{4(s_1(k-z_1))}}       \\
    \Leftrightarrow 1+e^{4(s_1(k-z_1))}                & = 1+e^{-4(s_0(k-z_0))}                \\
    \Leftrightarrow {4(s_1(k-z_1))}                    & = {-4(s_0(k-z_0))}                    \\
    \Leftrightarrow s_1k-s_1z_1                        & = -s_0k+s_0z_0                        \\
    \Leftrightarrow s_1k+s_0k                          & = s_0z_0 + s_1z_1                     \\
    \Leftrightarrow k                                  & = \frac{s_0z_0 + s_1z_1}{s_1+s_0}
\end{align*}
Given Equation 11,
\begin{align*}
    p_{c_t}^{(t,0)}(k)                                         & = p_{c_0}^{(t,0)}(k)     \\
    \Leftrightarrow 1 - \rho - p_{c_0}^{(t,0)}(k)              & = p_{c_0}^{(t,0)}(k)     \\
    \Leftrightarrow 1 - \rho                                   & = 2p_{c_0}^{(t,0)}(k)    \\
    \Leftrightarrow p_{c_0}^{(t,0)}(k)                         & = \nicefrac{(1-\rho)}{2} \\
    \Leftrightarrow p_{c_0}^{(t,0)}(k) \leq p_{c_t}^{(t,0)}(k) & \leq \nicefrac{1}{2}     \\
    \Rightarrow \max_{i\in[n]}f_i(x)                           & \leq \nicefrac{1}{2}
\end{align*}

\subsection{Binary search procedure} \label{binary_search}
All existing query-based attacks are based on performing a binary search between an image in the target class for targeted attacks or random noise for untargeted attacks and the source image in order to find the classification boundary.
The procedure crosses a low confidence region before outputting a boundary sample with a slightly higher probability for the target class. Subsequently, this image is used as a starting point for the exploration of the decision boundary of the query based attack that will then aim at minimizing the distortion of the original image.

\begin{figure*}[h]
    \label{catsanddogs}
    \centering
    \begin{subfigure}{.75\textwidth}
        \centering
        \includegraphics[trim=3cm 10.0cm 2.5cm 11.5cm,clip, width=1.0\linewidth]{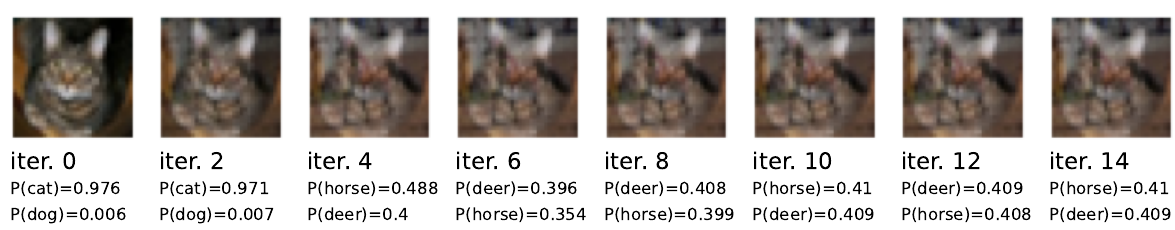}
    \end{subfigure}\\
    \begin{subfigure}{.75\textwidth}
        \centering
        \includegraphics[trim=3cm 10.0cm 2.5cm 11.5cm,clip, width=1.0\linewidth]{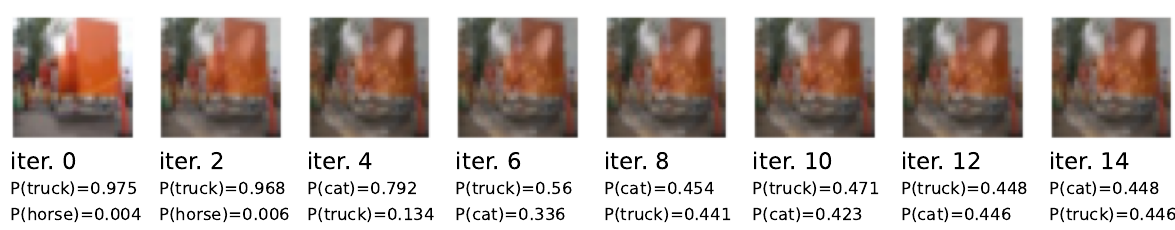}
    \end{subfigure}\\
    \begin{subfigure}{.75\textwidth}
        \centering
        \includegraphics[trim=3cm 10.0cm 2.5cm 11.5cm,clip, width=1.0\linewidth]{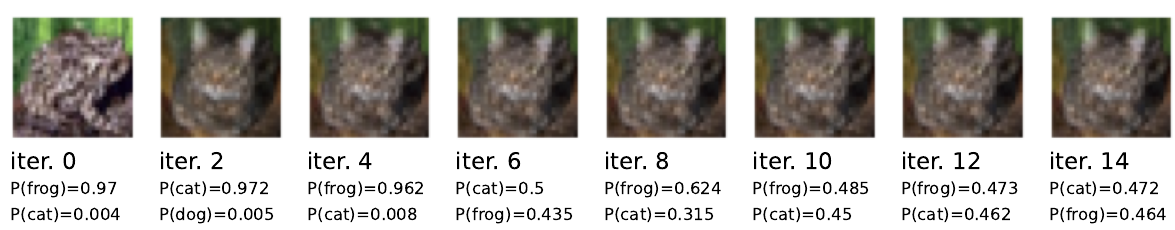}
    \end{subfigure}\\
    \begin{subfigure}{.75\textwidth}
        \centering
        \includegraphics[trim=3cm 10.0cm 2.5cm 11.5cm,clip, width=1.0\linewidth]{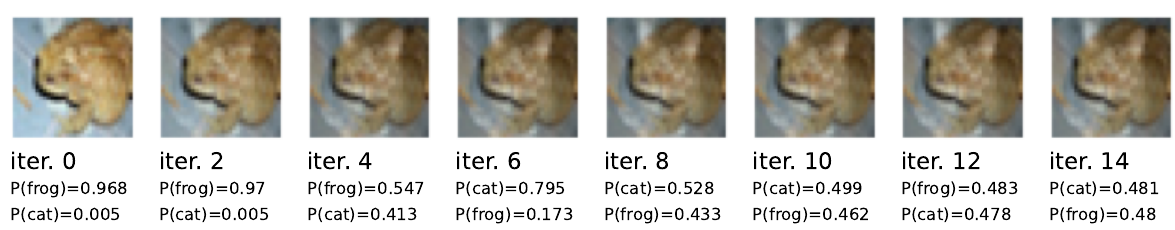}
    \end{subfigure}\\
    \begin{subfigure}{.75\textwidth}
        \centering
        \includegraphics[trim=3cm 10.0cm 2.5cm 11.5cm,clip, width=1.0\linewidth]{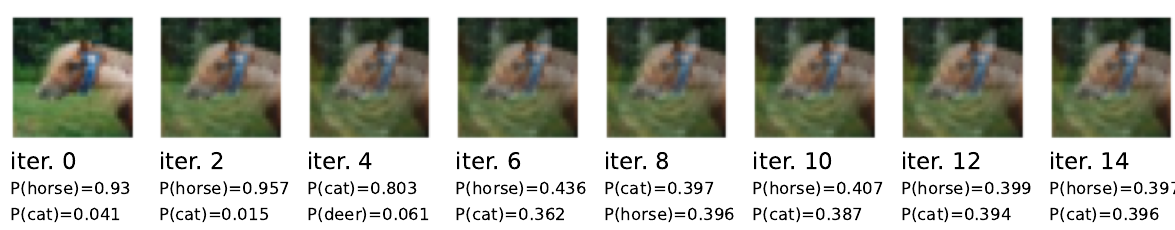}
    \end{subfigure}\\
    \begin{subfigure}{.75\textwidth}
        \centering
        \includegraphics[trim=3cm 10.0cm 2.5cm 11.5cm,clip, width=1.0\linewidth]{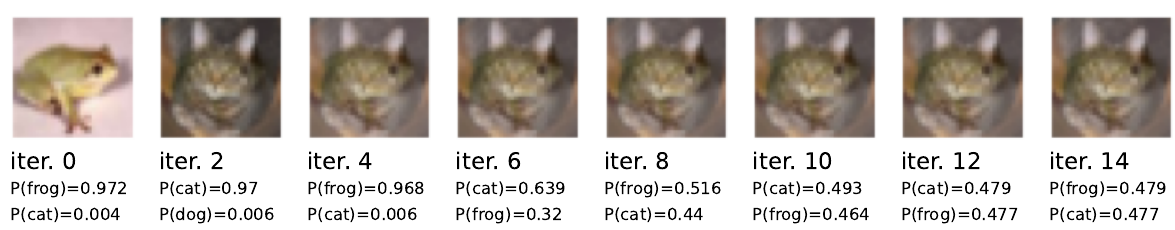}
    \end{subfigure}\\
    \begin{subfigure}{.75\textwidth}
        \centering
        \includegraphics[trim=3cm 10.0cm 2.5cm 11.5cm,clip, width=1.0\linewidth]{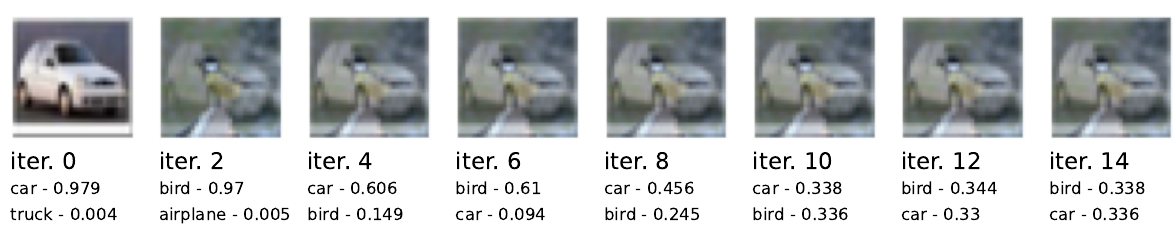}
    \end{subfigure}\\
    \begin{subfigure}{.75\textwidth}
        \centering
        \includegraphics[trim=3cm 10.0cm 2.5cm 11.5cm,clip, width=1.0\linewidth]{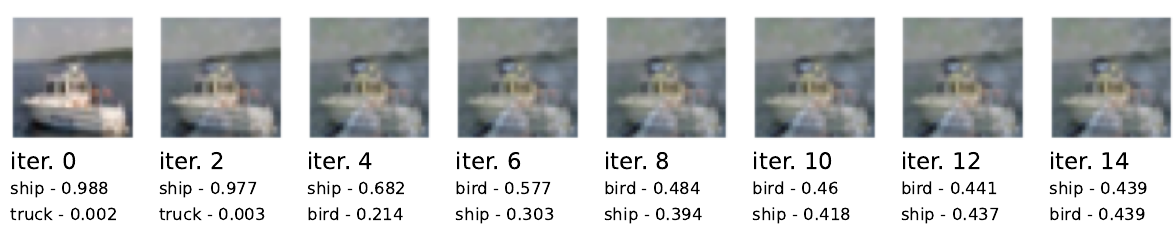}
    \end{subfigure}\\
    \caption{Selected iterations (or steps) from the binary search procedure typically used in a decision-based attack such as PSJA and SurFree. Here, a source image is blended with a target image. }
    \label{fig:catsdogs}
\end{figure*}

While the binary search process typically aims at finding the classification boundary between an original and some target class, the boundary does not necessarily need to be between the target and original classes. Indeed, since the images are highly distorted at this point of the process (recall that this is the first binary search of the full attack procedure), the model may classify images on the boundary between the target class and another class that is different from the original class. We depict this for several images in \Cref{fig:catsdogs} from the CIFAR-10 dataset for eight randomly chosen images that are blended into various target images.

\subsection{Attack choice}

\begin{figure}[H]
    \centering
    \includegraphics[width=1\columnwidth]{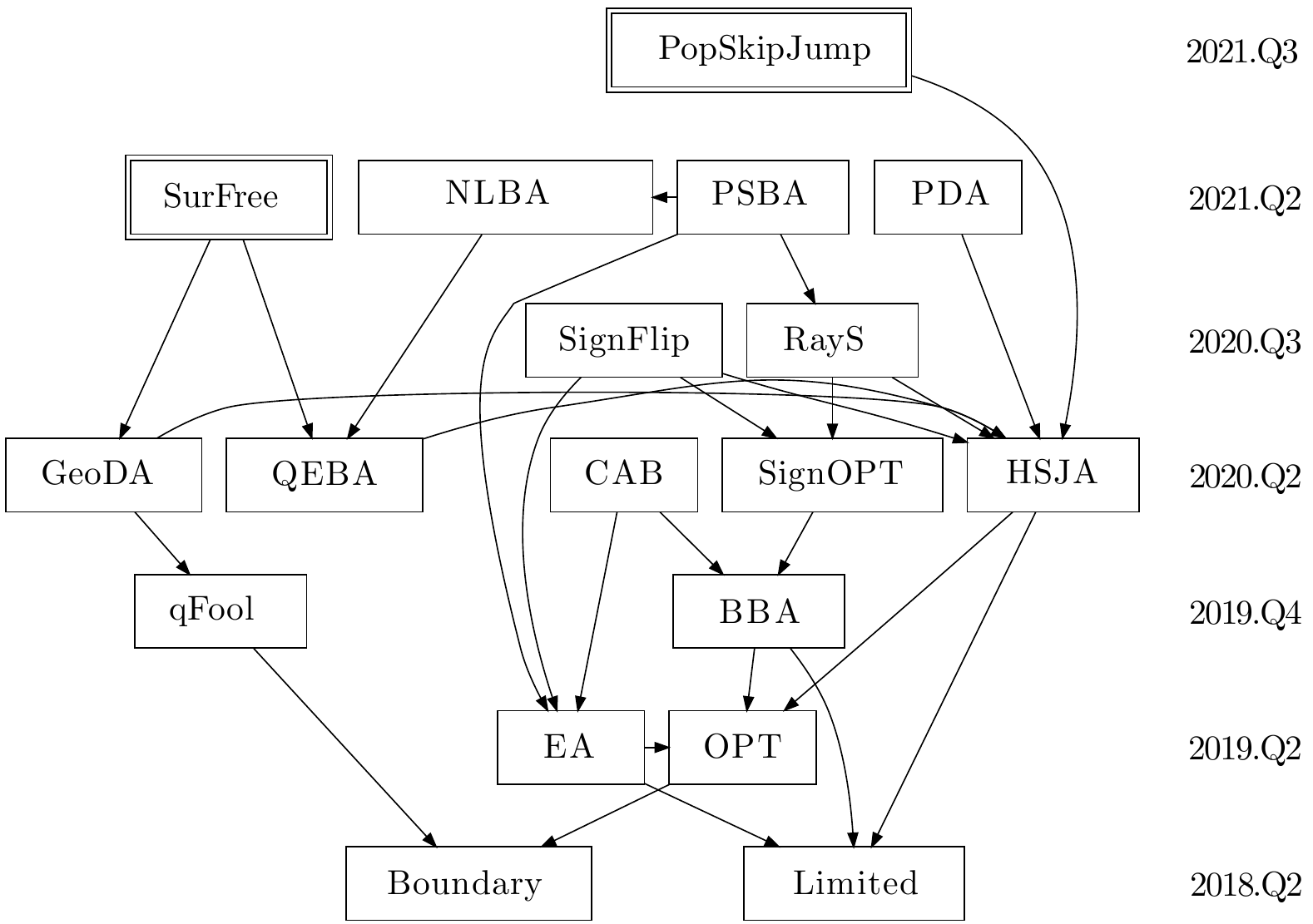}

    \caption{Overview of query-based attacks in the period between Q2 2018 and Q3 2021. Nodes refer to peer-reviewed contributions, and an arrow between paper X and paper Y (while transitive arrows have been omitted for clarity) refers to the fact that paper X evaluated its attack against the proposal in paper Y,  and reported an increase in adversarial success.}
    \label{fig:attack_choice}
\end{figure}

We thoroughly overview existing query-based attacks proposed between Q2 2018 and Q3 2021 and show them in \Cref{fig:attack_choice}. This enables us to extract the most powerful and representative query-based attacks and evaluate our approach against those, i.e., PopSkipJump \cite{simon-gabrielPopSkipJumpDecisionBasedAttack2021a} and SurFree \cite{mahoSurFreeFastSurrogatefree2021a}. The abbreviations are mapped to the references as follows:
\begin{itemize}
    \item Boundary \cite{brendelDecisionbasedAdversarialAttacks2018a}
    \item Limited \cite{ilyasBlackboxAdversarialAttacksa}
    \item EA \cite{dongEfficientDecisionBasedBlackBox2019a}
    \item OPT \cite{chengQueryEfficientHardLabelBlackBox2019a}
    \item qFool \cite{liuGeometryInspiredDecisionBasedAttack2019a}
    \item BBA \cite{brunnerGuessingSmartBiased2019a}
    \item GeoDA \cite{rahmatiGeoDAGeometricFramework2020a}
    \item QEBA \cite{liQEBAQueryEfficientBoundaryBased2020a}
    \item CAB \cite{shiPolishingDecisionBasedAdversarial2020a}
    \item SignOPT \cite{chengSignOPTQueryEfficientHardlabel2020a}
    \item HSJA \cite{chenHopSkipJumpAttackQueryEfficientDecisionBased2020a}
    \item SignFlip \cite{chenBoostingDecisionBasedBlackBox2020a}
    \item RayS \cite{chenRaySRaySearching2020a}
    \item SurFree \cite{mahoSurFreeFastSurrogatefree2021a}
    \item NLBA \cite{liNonlinearProjectionBased2021a}
    \item PSBA \cite{zhangProgressiveScaleBoundaryBlackbox2021a}
    \item PDA \cite{yanPolicyDrivenAttackLearning2021a}
    \item PopSkipJump \cite{simon-gabrielPopSkipJumpDecisionBasedAttack2021a}
\end{itemize}

\subsection{Full results}
\Cref{tab:full_results} contains the complete evaluation results with all possible combinations of $\tau$ and $\nu$ for PSJA and SurFree on the CIFAR-10, CIFAR-100, and ImageNet dataset.

\subsubsection{ImageNet} \label{imagenet_data}

We notice that the $\robustacc$ decreases with increasing the parameters $\nu$ and $\tau$. This is related to the vastly larger input space and the higher number of classes, for which introduced noise can result in easier label flips, which actually boosts the success of an adversary in the untargeted attack setting.

\begin{table*}[t]
    \centering
    \resizebox{1.00\textwidth}{!}{%
        \begin{tabular}{cc|c|cccccc|ccccccc|ccccccc}
            \toprule
             & Dataset        &                 & \multicolumn{6}{c|}{PSJA with RND} & \multicolumn{7}{c|}{PSJA with RCR} & \multicolumn{7}{c}{SurFree with JPEG}                                                                     \\
             &                &                 &                                    &                                    &                                       &      &      &     &         &  &  &  &  &  &  &  &  &  &  &  &  & \\
             & $\nu$ / $\tau$ &                 & 0.01                               & 0.02                               & 0.05                                  & 0.07 & 0.08 & 0.1 & \shs{29                                       \\200} & \shs{27\\176} & \shs{26\\152} & \shs{25\\128} & \shs{22\\104} & \shs{18\\80} & \shs{14\\56}  & 85 & 75 & 60 & 50 & 35 & 25 & 10 \\
            \midrule

\parbox[t]{2mm}{\multirow{22}{*}{\rotatebox[origin=c]{90}{CIFAR-10}}}
& \shs{0.0} &\shs{$\mainacc$\\$\robustacc$} &\shs{0.95\\0.47}&\shs{0.95\\0.47}&\shs{0.95\\0.47}&\shs{0.95\\0.47}&\shs{0.95\\0.47}&\shs{0.95\\0.47}&\shs{0.95\\0.41}&\shs{0.95\\0.41}&\shs{0.95\\0.41}&\shs{0.95\\0.41}&\shs{0.95\\0.41}&\shs{0.95\\0.41}&\shs{0.95\\0.41}&\shs{0.95\\0.21}&\shs{0.95\\0.21}&\shs{0.95\\0.21}&\shs{0.95\\0.21}&\shs{0.95\\0.21}&\shs{0.95\\0.21}&\shs{0.95\\0.21}\\\addlinespace
& \shs{0.3} &\shs{$\mainacc$\\$\robustacc$} &\shs{0.95\\0.44}&\shs{0.95\\0.42}&\shp{0.95\\0.49}&\shs{0.95\\0.47}&\shp{0.95\\0.48}&\shs{0.95\\0.45}&\shp{0.95\\0.44}&\shs{0.95\\0.41}&\shp{0.95\\0.47}&\shp{0.95\\0.46}&\shp{0.95\\0.45}&\shp{0.95\\0.49}&\shp{0.95\\0.49}&\shp{0.95\\0.26}&\shs{0.95\\0.21}&\shp{0.95\\0.42}&\shs{0.95\\0.19}&\shp{0.95\\0.28}&\shp{0.95\\0.39}&\shs{0.95\\0.17}\\\addlinespace
& \shs{0.4} &\shs{$\mainacc$\\$\robustacc$} &\shp{0.95\\0.52}&\shs{0.95\\0.46}&\shs{0.95\\0.43}&\shs{0.95\\0.45}&\shs{0.95\\0.46}&\shp{0.95\\0.51}&\shp{0.95\\0.42}&\shs{0.95\\0.39}&\shs{0.95\\0.39}&\shs{0.95\\0.40}&\shp{0.95\\0.42}&\shp{0.95\\0.47}&\shp{0.95\\0.45}&\shp{0.95\\0.33}&\shp{0.95\\0.31}&\shp{0.95\\0.23}&\shp{0.95\\0.30}&\shp{0.95\\0.32}&\shp{0.95\\0.47}&\shp{0.95\\0.51}\\\addlinespace
& \shs{0.5} &\shs{$\mainacc$\\$\robustacc$} &\shp{0.95\\0.49}&\shp{0.95\\0.51}&\shs{0.95\\0.44}&\shs{0.95\\0.44}&\shs{0.94\\0.48}&\shs{0.94\\0.47}&\shs{0.95\\0.37}&\shp{0.95\\0.43}&\shs{0.95\\0.38}&\shs{0.95\\0.41}&\shp{0.95\\0.42}&\shp{0.95\\0.48}&\shp{0.95\\0.47}&\shp{0.95\\0.52}&\shp{0.95\\0.37}&\shp{0.95\\0.41}&\shp{0.95\\0.55}&\shp{0.95\\0.47}&\shp{0.95\\0.30}&\shp{0.95\\0.26}\\\addlinespace
& \shs{0.6} &\shs{$\mainacc$\\$\robustacc$} &\shp{0.95\\0.50}&\shp{0.95\\0.54}&\shs{0.94\\0.49}&\shp{0.94\\0.54}&\shs{0.94\\0.50}&\shs{0.94\\0.50}&\shp{0.95\\0.47}&\shs{0.95\\0.39}&\shs{0.95\\0.41}&\shs{0.95\\0.35}&\shp{0.94\\0.42}&\shp{0.94\\0.49}&\shp{0.94\\0.43}&\shp{0.95\\0.36}&\shp{0.94\\0.57}&\shp{0.94\\0.50}&\shp{0.94\\0.46}&\shs{0.94\\0.34}&\shs{0.94\\0.33}&\shp{0.94\\0.43}\\\addlinespace
& \shs{0.7} &\shs{$\mainacc$\\$\robustacc$} &\shs{0.94\\0.49}&\shp{0.94\\0.57}&\shs{0.94\\0.47}&\shs{0.94\\0.48}&\shs{0.94\\0.48}&\shs{0.93\\0.49}&\shp{0.94\\0.44}&\shp{0.94\\0.41}&\shp{0.94\\0.45}&\shp{0.94\\0.43}&\shp{0.94\\0.45}&\shp{0.94\\0.46}&\shp{0.94\\0.43}&\shp{0.94\\0.40}&\shp{0.94\\0.49}&\shs{0.94\\0.31}&\shs{0.94\\0.32}&\shs{0.94\\0.26}&\shp{0.94\\0.41}&\shp{0.94\\0.40}\\\addlinespace
& \shs{0.8} &\shs{$\mainacc$\\$\robustacc$} &\shp{0.95\\0.55}&\shp{0.94\\0.57}&\shp{0.94\\0.59}&\shs{0.93\\0.51}&\shs{0.93\\0.43}&\shs{0.93\\0.50}&\shs{0.94\\0.35}&\shp{0.94\\0.41}&\shs{0.94\\0.39}&\shs{0.94\\0.39}&\shp{0.94\\0.43}&\shp{0.93\\0.46}&\shp{0.93\\0.45}&\shp{0.94\\0.67}&\shp{0.94\\0.40}&\shp{0.94\\0.44}&\shs{0.94\\0.38}&\shs{0.93\\0.33}&\shp{0.94\\0.54}&\shs{0.93\\0.42}\\\addlinespace
& \shs{0.9} &\shs{$\mainacc$\\$\robustacc$} &\shs{0.94\\0.49}&\shp{0.94\\0.56}&\shp{0.93\\0.54}&\shp{0.92\\0.54}&\shs{0.91\\0.48}&\shs{0.91\\0.49}&\shp{0.94\\0.42}&\shs{0.94\\0.38}&\shp{0.93\\0.47}&\shs{0.93\\0.35}&\shp{0.93\\0.45}&\shs{0.92\\0.37}&\shp{0.91\\0.43}&\shp{0.94\\0.44}&\shp{0.93\\0.61}&\shs{0.93\\0.36}&\shs{0.93\\0.55}&\shs{0.92\\0.60}&\shp{0.93\\0.56}&\shs{0.92\\0.33}\\\addlinespace
& \shs{0.97} &\shs{$\mainacc$\\$\robustacc$} &\shs{0.95\\0.46}&\shp{0.94\\0.62}&\shp{0.87\\0.64}&\shs{0.82\\0.54}&\shs{0.78\\0.47}&\shs{0.73\\0.47}&\shs{0.92\\0.32}&\shp{0.90\\0.42}&\shs{0.89\\0.33}&\shp{0.88\\0.46}&\shp{0.81\\0.43}&\shp{0.72\\0.48}&\shs{0.66\\0.37}&\shp{0.92\\0.83}&\shp{0.91\\0.84}&\shp{0.90\\0.77}&\shp{0.88\\0.75}&\shp{0.86\\0.75}&\shp{0.84\\0.74}&\shs{0.73\\0.71}\\\addlinespace
& \shs{0.99} &\shs{$\mainacc$\\$\robustacc$} &\shs{0.94\\0.51}&\shp{0.94\\0.55}&\shp{0.83\\0.71}&\shs{0.65\\0.58}&\shs{0.55\\0.53}&\shs{0.37\\0.35}&\shs{0.91\\0.33}&\shs{0.88\\0.39}&\shs{0.86\\0.34}&\shs{0.83\\0.34}&\shp{0.66\\0.51}&\shs{0.39\\0.39}&\shs{0.21\\0.25}&\shs{0.92\\0.65}&\shs{0.90\\0.71}&\shs{0.87\\0.60}&\shs{0.85\\0.60}&\shs{0.80\\0.57}&\shs{0.75\\0.59}&\shs{0.48\\0.49}\\\addlinespace
& \shs{1.0} &\shs{$\mainacc$\\$\robustacc$} &\shs{0.94\\0.48}&\shs{0.94\\0.53}&\shs{0.82\\0.71}&\shs{0.65\\0.60}&\shs{0.54\\0.50}&\shs{0.36\\0.33}&\shs{0.91\\0.37}&\shs{0.88\\0.34}&\shs{0.86\\0.39}&\shs{0.83\\0.33}&\shs{0.67\\0.49}&\shs{0.38\\0.41}&\shs{0.21\\0.23}&\shs{0.92\\0.73}&\shs{0.90\\0.67}&\shs{0.87\\0.60}&\shs{0.85\\0.64}&\shs{0.80\\0.63}&\shs{0.75\\0.59}&\shs{0.48\\0.50}\\\midrule\parbox[t]{2mm}{\multirow{22}{*}{\rotatebox[origin=c]{90}{CIFAR-100}}}
& \shs{0.0} &\shs{$\mainacc$\\$\robustacc$} &\shs{0.60\\0.15}&\shs{0.60\\0.15}&\shs{0.60\\0.15}&\shs{0.60\\0.15}&\shs{0.60\\0.15}&\shs{0.60\\0.15}&\shs{0.60\\0.17}&\shs{0.60\\0.17}&\shs{0.60\\0.17}&\shs{0.60\\0.17}&\shs{0.60\\0.17}&\shs{0.60\\0.17}&\shs{0.60\\0.17}&\shs{0.60\\0.52}&\shs{0.60\\0.52}&\shs{0.60\\0.52}&\shs{0.60\\0.52}&\shs{0.60\\0.52}&\shs{0.60\\0.52}&\shs{0.60\\0.52}\\\addlinespace
& \shs{0.3} &\shs{$\mainacc$\\$\robustacc$} &\shs{0.60\\0.17}&\shs{0.60\\0.17}&\shs{0.59\\0.14}&\shs{0.59\\0.13}&\shs{0.59\\0.19}&\shs{0.59\\0.15}&\shs{0.60\\0.16}&\shs{0.60\\0.15}&\shs{0.60\\0.13}&\shs{0.60\\0.15}&\shs{0.60\\0.13}&\shs{0.59\\0.15}&\shs{0.58\\0.14}&\shs{0.60\\0.33}&\shs{0.60\\0.36}&\shp{0.60\\0.65}&\shp{0.60\\0.73}&\shs{0.60\\0.43}&\shs{0.60\\0.33}&\shp{0.59\\0.73}\\\addlinespace
& \shs{0.4} &\shs{$\mainacc$\\$\robustacc$} &\shs{0.60\\0.17}&\shs{0.60\\0.18}&\shs{0.59\\0.14}&\shs{0.58\\0.13}&\shs{0.58\\0.20}&\shs{0.57\\0.18}&\shp{0.60\\0.18}&\shs{0.59\\0.17}&\shp{0.59\\0.21}&\shp{0.59\\0.21}&\shp{0.59\\0.21}&\shs{0.57\\0.16}&\shs{0.56\\0.16}&\shs{0.59\\0.43}&\shs{0.59\\0.43}&\shs{0.59\\0.53}&\shp{0.59\\0.80}&\shp{0.59\\0.75}&\shs{0.58\\0.66}&\shs{0.57\\0.72}\\\addlinespace
& \shs{0.5} &\shs{$\mainacc$\\$\robustacc$} &\shs{0.60\\0.17}&\shp{0.59\\0.26}&\shs{0.58\\0.23}&\shs{0.56\\0.22}&\shs{0.56\\0.19}&\shs{0.55\\0.18}&\shp{0.59\\0.24}&\shp{0.59\\0.25}&\shs{0.58\\0.21}&\shp{0.58\\0.27}&\shs{0.57\\0.24}&\shs{0.55\\0.23}&\shs{0.53\\0.13}&\shp{0.59\\0.79}&\shs{0.59\\0.45}&\shs{0.58\\0.65}&\shs{0.58\\0.49}&\shs{0.58\\0.47}&\shs{0.57\\0.63}&\shs{0.55\\0.74}\\\addlinespace
& \shs{0.6} &\shs{$\mainacc$\\$\robustacc$} &\shs{0.60\\0.21}&\shp{0.59\\0.29}&\shs{0.56\\0.26}&\shs{0.54\\0.20}&\shs{0.54\\0.29}&\shs{0.52\\0.29}&\shp{0.59\\0.29}&\shp{0.58\\0.37}&\shp{0.58\\0.37}&\shp{0.57\\0.28}&\shs{0.55\\0.27}&\shs{0.52\\0.28}&\shs{0.50\\0.21}&\shs{0.58\\0.47}&\shp{0.58\\0.78}&\shs{0.57\\0.52}&\shp{0.57\\0.82}&\shs{0.56\\0.51}&\shs{0.55\\0.77}&\shs{0.53\\0.43}\\\addlinespace
& \shs{0.7} &\shs{$\mainacc$\\$\robustacc$} &\shs{0.60\\0.20}&\shp{0.59\\0.28}&\shp{0.55\\0.31}&\shs{0.53\\0.28}&\shs{0.51\\0.28}&\shs{0.49\\0.33}&\shp{0.58\\0.31}&\shp{0.57\\0.40}&\shp{0.56\\0.36}&\shp{0.56\\0.34}&\shs{0.53\\0.29}&\shs{0.50\\0.34}&\shs{0.47\\0.22}&\shp{0.58\\0.90}&\shp{0.57\\0.81}&\shp{0.56\\0.89}&\shs{0.56\\0.56}&\shs{0.55\\0.56}&\shs{0.54\\0.70}&\shs{0.50\\0.69}\\\addlinespace
& \shs{0.8} &\shs{$\mainacc$\\$\robustacc$} &\shp{0.60\\0.25}&\shp{0.59\\0.33}&\shp{0.53\\0.38}&\shs{0.49\\0.35}&\shs{0.48\\0.30}&\shs{0.45\\0.28}&\shp{0.56\\0.37}&\shp{0.56\\0.42}&\shp{0.54\\0.39}&\shp{0.54\\0.44}&\shs{0.50\\0.36}&\shs{0.46\\0.28}&\shs{0.42\\0.18}&\shs{0.57\\0.51}&\shs{0.56\\0.50}&\shs{0.55\\0.60}&\shp{0.54\\0.88}&\shs{0.53\\0.57}&\shs{0.52\\0.52}&\shs{0.46\\0.79}\\\addlinespace
& \shs{0.9} &\shs{$\mainacc$\\$\robustacc$} &\shp{0.60\\0.25}&\shp{0.59\\0.31}&\shs{0.50\\0.33}&\shs{0.46\\0.37}&\shs{0.43\\0.34}&\shs{0.40\\0.40}&\shp{0.55\\0.42}&\shp{0.53\\0.43}&\shp{0.52\\0.49}&\shs{0.50\\0.42}&\shs{0.46\\0.40}&\shs{0.40\\0.29}&\shs{0.34\\0.25}&\shp{0.56\\0.94}&\shs{0.56\\0.51}&\shs{0.53\\0.50}&\shs{0.52\\0.52}&\shs{0.51\\0.60}&\shs{0.49\\0.58}&\shs{0.41\\0.78}\\\addlinespace
& \shs{0.97} &\shs{$\mainacc$\\$\robustacc$} &\shs{0.60\\0.20}&\shp{0.58\\0.34}&\shs{0.48\\0.38}&\shs{0.41\\0.37}&\shs{0.38\\0.41}&\shs{0.33\\0.39}&\shp{0.54\\0.36}&\shs{0.51\\0.41}&\shp{0.49\\0.47}&\shs{0.47\\0.45}&\shs{0.39\\0.37}&\shs{0.30\\0.35}&\shs{0.23\\0.22}&\shs{0.56\\0.85}&\shs{0.55\\0.60}&\shs{0.52\\0.50}&\shs{0.51\\0.56}&\shp{0.49\\0.96}&\shp{0.45\\0.93}&\shs{0.35\\0.57}\\\addlinespace
& \shs{0.99} &\shs{$\mainacc$\\$\robustacc$} &\shs{0.60\\0.19}&\shp{0.58\\0.33}&\shs{0.47\\0.40}&\shs{0.38\\0.44}&\shs{0.35\\0.36}&\shs{0.28\\0.36}&\shp{0.54\\0.44}&\shp{0.50\\0.49}&\shp{0.47\\0.50}&\shs{0.45\\0.47}&\shs{0.36\\0.43}&\shs{0.25\\0.33}&\shs{0.16\\0.18}&\shs{0.56\\0.49}&\shp{0.54\\0.89}&\shs{0.52\\0.55}&\shs{0.50\\0.60}&\shs{0.48\\0.58}&\shs{0.44\\0.50}&\shp{0.31\\0.91}\\\addlinespace
& \shs{1.0} &\shs{$\mainacc$\\$\robustacc$} &\shs{0.60\\0.24}&\shs{0.58\\0.26}&\shs{0.46\\0.46}&\shs{0.37\\0.37}&\shs{0.34\\0.42}&\shs{0.27\\0.37}&\shs{0.53\\0.35}&\shs{0.50\\0.46}&\shs{0.47\\0.48}&\shs{0.44\\0.47}&\shs{0.34\\0.43}&\shs{0.22\\0.31}&\shs{0.12\\0.21}&\shs{0.56\\0.85}&\shs{0.54\\0.82}&\shs{0.52\\0.83}&\shs{0.50\\0.83}&\shs{0.47\\0.76}&\shs{0.43\\0.50}&\shs{0.29\\0.35}\\\midrule\parbox[t]{2mm}{\multirow{22}{*}{\rotatebox[origin=c]{90}{ImageNet}}}
& \shs{0.0} &\shs{$\mainacc$\\$\robustacc$} &\shs{0.81\\0.97}&\shs{0.81\\0.97}&\shs{0.81\\0.97}&\shs{0.81\\0.97}&\shs{0.81\\0.97}&\shs{0.81\\0.97}&\shs{0.81\\0.97}&\shs{0.81\\0.97}&\shs{0.81\\0.97}&\shs{0.81\\0.97}&\shs{0.81\\0.97}&\shs{0.81\\0.97}&\shs{0.81\\0.97}&\shs{0.81\\0.61}&\shs{0.81\\0.61}&\shs{0.81\\0.61}&\shs{0.81\\0.61}&\shs{0.81\\0.61}&\shs{0.81\\0.61}&\shs{0.81\\0.61}\\\addlinespace
& \shs{0.3} &\shs{$\mainacc$\\$\robustacc$} &\shs{0.81\\0.97}&\shs{0.81\\0.97}&\shs{0.80\\0.96}&\shs{0.80\\0.96}&\shs{0.80\\0.95}&\shs{0.80\\0.94}&\shs{0.81\\0.96}&\shs{0.80\\0.96}&\shs{0.80\\0.94}&\shs{0.80\\0.95}&\shs{0.79\\0.94}&\shs{0.79\\0.94}&\shs{0.78\\0.94}&\shp{0.80\\0.77}&\shp{0.80\\0.82}&\shp{0.80\\0.80}&\shp{0.80\\0.80}&\shp{0.80\\0.76}&\shp{0.80\\0.79}&\shp{0.79\\0.76}\\\addlinespace
& \shs{0.4} &\shs{$\mainacc$\\$\robustacc$} &\shs{0.81\\0.97}&\shs{0.80\\0.96}&\shs{0.80\\0.94}&\shs{0.79\\0.94}&\shs{0.79\\0.94}&\shs{0.78\\0.95}&\shs{0.80\\0.95}&\shs{0.80\\0.93}&\shs{0.80\\0.94}&\shs{0.79\\0.93}&\shs{0.78\\0.92}&\shs{0.77\\0.93}&\shs{0.76\\0.92}&\shp{0.80\\0.80}&\shp{0.80\\0.79}&\shp{0.79\\0.79}&\shp{0.79\\0.77}&\shp{0.79\\0.79}&\shp{0.79\\0.75}&\shs{0.78\\0.74}\\\addlinespace
& \shs{0.5} &\shs{$\mainacc$\\$\robustacc$} &\shp{0.80\\0.98}&\shp{0.80\\0.98}&\shs{0.79\\0.94}&\shs{0.78\\0.95}&\shs{0.78\\0.95}&\shs{0.77\\0.92}&\shs{0.80\\0.96}&\shs{0.80\\0.93}&\shs{0.79\\0.92}&\shs{0.78\\0.95}&\shs{0.76\\0.91}&\shs{0.74\\0.91}&\shs{0.72\\0.89}&\shp{0.79\\0.79}&\shp{0.79\\0.79}&\shp{0.79\\0.76}&\shs{0.78\\0.75}&\shs{0.78\\0.73}&\shs{0.77\\0.74}&\shs{0.75\\0.70}\\\addlinespace
& \shs{0.6} &\shs{$\mainacc$\\$\robustacc$} &\shs{0.80\\0.97}&\shp{0.80\\0.98}&\shs{0.78\\0.94}&\shs{0.77\\0.94}&\shs{0.76\\0.92}&\shs{0.75\\0.90}&\shp{0.80\\0.97}&\shs{0.79\\0.95}&\shs{0.78\\0.91}&\shs{0.76\\0.93}&\shs{0.74\\0.93}&\shs{0.71\\0.88}&\shs{0.67\\0.82}&\shp{0.79\\0.79}&\shs{0.78\\0.76}&\shs{0.78\\0.76}&\shs{0.77\\0.75}&\shs{0.77\\0.72}&\shs{0.76\\0.71}&\shs{0.73\\0.70}\\\addlinespace
& \shs{0.7} &\shs{$\mainacc$\\$\robustacc$} &\shs{0.80\\0.97}&\shs{0.80\\0.97}&\shs{0.77\\0.90}&\shs{0.76\\0.89}&\shs{0.75\\0.87}&\shs{0.73\\0.86}&\shs{0.80\\0.95}&\shs{0.79\\0.93}&\shs{0.77\\0.93}&\shs{0.75\\0.92}&\shs{0.71\\0.88}&\shs{0.66\\0.85}&\shs{0.60\\0.75}&\shp{0.79\\0.77}&\shs{0.77\\0.75}&\shs{0.76\\0.74}&\shs{0.76\\0.75}&\shs{0.75\\0.73}&\shs{0.74\\0.68}&\shs{0.69\\0.64}\\\addlinespace
& \shs{0.8} &\shs{$\mainacc$\\$\robustacc$} &\shs{0.80\\0.97}&\shs{0.80\\0.97}&\shs{0.77\\0.91}&\shs{0.75\\0.88}&\shs{0.74\\0.88}&\shs{0.71\\0.81}&\shs{0.80\\0.94}&\shs{0.78\\0.93}&\shs{0.77\\0.93}&\shs{0.73\\0.89}&\shs{0.67\\0.88}&\shs{0.58\\0.78}&\shs{0.47\\0.58}&\shs{0.78\\0.76}&\shs{0.77\\0.73}&\shs{0.75\\0.68}&\shs{0.74\\0.69}&\shs{0.73\\0.69}&\shs{0.71\\0.64}&\shs{0.64\\0.60}\\\addlinespace
& \shs{0.9} &\shs{$\mainacc$\\$\robustacc$} &\shp{0.80\\0.98}&\shs{0.80\\0.97}&\shs{0.77\\0.91}&\shs{0.74\\0.88}&\shs{0.72\\0.85}&\shs{0.69\\0.82}&\shp{0.80\\0.97}&\shs{0.78\\0.94}&\shs{0.76\\0.93}&\shs{0.71\\0.90}&\shs{0.61\\0.81}&\shs{0.48\\0.70}&\shs{0.29\\0.39}&\shs{0.78\\0.78}&\shs{0.76\\0.70}&\shs{0.74\\0.65}&\shs{0.73\\0.68}&\shs{0.71\\0.66}&\shs{0.69\\0.61}&\shs{0.58\\0.51}\\\addlinespace
& \shs{0.97} &\shs{$\mainacc$\\$\robustacc$} &\shs{0.80\\0.97}&\shs{0.80\\0.97}&\shs{0.76\\0.92}&\shs{0.74\\0.89}&\shs{0.72\\0.84}&\shs{0.68\\0.79}&\shs{0.80\\0.96}&\shs{0.78\\0.94}&\shs{0.76\\0.93}&\shs{0.70\\0.92}&\shs{0.60\\0.80}&\shs{0.44\\0.67}&\shs{0.23\\0.40}&\shs{0.78\\0.78}&\shs{0.76\\0.72}&\shs{0.74\\0.64}&\shs{0.73\\0.67}&\shs{0.71\\0.66}&\shs{0.68\\0.62}&\shs{0.56\\0.48}\\\addlinespace
& \shs{0.99} &\shs{$\mainacc$\\$\robustacc$} &\shs{0.80\\0.97}&\shs{0.80\\0.97}&\shs{0.76\\0.91}&\shs{0.74\\0.85}&\shs{0.72\\0.85}&\shs{0.68\\0.78}&\shs{0.80\\0.95}&\shs{0.78\\0.95}&\shs{0.75\\0.91}&\shs{0.70\\0.90}&\shs{0.59\\0.84}&\shs{0.43\\0.68}&\shs{0.23\\0.36}&\shs{0.78\\0.78}&\shs{0.76\\0.72}&\shs{0.74\\0.64}&\shs{0.73\\0.67}&\shs{0.71\\0.66}&\shs{0.68\\0.62}&\shs{0.56\\0.48}\\\addlinespace
& \shs{1.0} &\shs{$\mainacc$\\$\robustacc$} &\shs{0.80\\0.97}&\shs{0.80\\0.97}&\shs{0.76\\0.91}&\shs{0.74\\0.85}&\shs{0.72\\0.84}&\shs{0.68\\0.79}&\shs{0.80\\0.95}&\shs{0.78\\0.94}&\shs{0.76\\0.91}&\shs{0.70\\0.91}&\shs{0.59\\0.83}&\shs{0.43\\0.71}&\shs{0.22\\0.33}&\shs{0.78\\0.78}&\shs{0.76\\0.72}&\shs{0.74\\0.64}&\shs{0.73\\0.67}&\shs{0.71\\0.66}&\shs{0.68\\0.62}&\shs{0.56\\0.48}\\\bottomrule

        \end{tabular}
    }
    \caption{$\robustacc$ and $\mainacc$ based on the defense parameter $\nu$ and the threshold $\tau$. We mark in \textbf{bold} the improvements beyond the Pareto frontier achieved in vanilla constructs (i.e., when $\tau=1.0$). }
    \label{tab:full_results}
\end{table*}

\end{document}